\documentclass[journal]{IEEEtran}

\usepackage[OT1]{fontenc} 
\usepackage{array}
\usepackage{textcomp}
\usepackage{color}
\usepackage{colortbl}
\usepackage{diagbox} 
\usepackage{graphicx}
\usepackage{amsmath}
\usepackage{algorithm}
\usepackage{algpseudocode}
\usepackage{subfigure}
\usepackage{stfloats}
\usepackage{url}
\usepackage{multirow}
\usepackage{multicol}
\usepackage[table,xcdraw]{xcolor}
\usepackage{amssymb}
\usepackage{booktabs}
\usepackage{arydshln}
\usepackage{slashbox}
\usepackage{makecell}
\usepackage{rotating}
\usepackage{tabularx}
\usepackage{threeparttable}
\usepackage[colorlinks=true,linkcolor=blue]{hyperref}
\usepackage{pifont}
\usepackage{amssymb}
\usepackage{mathrsfs}

\usepackage[hyperref=true,style=ieee,backend=biber,sorting=none,maxnames=1,minnames=1]{biblatex}
\begin{document}
\title{\LARGE \bf
	An Active and Contrastive Learning Framework for Fine-Grained Off-Road Semantic Segmentation}

\author{Biao~Gao$^{1}$,~\IEEEmembership{Member,~IEEE,}
		Xijun~Zhao$^{2}$,~\IEEEmembership{Member,~IEEE,}
		Huijing~Zhao$^{1}$,~\IEEEmembership{Member,~IEEE,}
	\thanks{*This work was supported in part by the National Natural Science Foundation of China under Grant 61973004 and High-performance Computing Platform of Peking University.}
	\thanks{$^{1}$B. Gao, and H. Zhao are with the Key Lab of Machine Perception (MOE), Peking University, Beijing, China. $^{2}$X. Zhao is with China North Vehicle Research Institute, Beijing, China.}%
	\thanks{Correspondence: H. Zhao, {\tt\small zhaohj@cis.pku.edu.cn}.}%

}

\markboth{IEEE TRANSACTIONS ON INTELLIGENT TRANSPORTATION SYSTEMS,~Vol.~?, No.~?, ??~????}%
{Shell \MakeLowercase{\textit{et al.}}: IEEE TRANSACTIONS ON INTELLIGENT TRANSPORTATION SYSTEMS}

\maketitle

\begin{abstract}
Off-road semantic segmentation with fine-grained labels is necessary for autonomous vehicles to understand driving scenes, as the coarse-grained road detection can not satisfy off-road vehicles with various mechanical properties.
Fine-grained semantic segmentation in off-road scenes usually has no unified category definition due to ambiguous nature environments, and the cost of pixel-wise labeling is extremely high.
Furthermore, semantic properties of off-road scenes can be very changeable due to various precipitations, temperature, defoliation, etc.
To address these challenges, this research proposes an active and contrastive learning-based method that does not rely on pixel-wise labels, but only on patch-based weak annotations for model learning.
There is no need for predefined semantic categories, the contrastive learning-based feature representation and adaptive clustering will discover the category model from scene data.
In order to actively adapt to new scenes, a risk evaluation method is proposed to discover and select hard frames with high-risk predictions for supplemental labeling, so as to update the model efficiently.
Experiments conducted on our self-developed off-road dataset and DeepScene dataset demonstrate that fine-grained semantic segmentation can be learned with only dozens of weakly labeled frames, and the model can efficiently adapt across scenes by weak supervision, while achieving almost the same level of performance as typical fully supervised baselines.

\end{abstract}


\section{Introduction} \label{sec:1}

\IEEEPARstart{S}{emantic} segmentation is one of the key perception techniques for an autonomous driving agent to navigate safely and smoothly in complex environments~\cite{badue2021self}.
There has been a large body of studies on semantic segmentation, while most of them are addressed in structured urban scenes~\cite{siam2017deep}.
Such scenes are composed of many man-made objects such as paved roads, lane markings, traffic signals, buildings, etc. These objects belong to semantically interpretable categories and their data have fairly clear boundaries.
Despite the large needs of fine-grained perception for autonomous driving at off-road scenes~\cite{braid2006terramax,shariati2019towards,fang2006trajectory}, semantic segmentation in such scenes has far less been studied. Off-road scenes are composed of natural objects in various shapes and of indistinct semantic category, diverse terrain surfaces, and changed topographical conditions~\cite{mei2017scene}. Semantic segmentation in such scenes remains an open challenge.

According to the granularity of scene understanding, the methods of off-road semantic segmentation can be divided into two groups: \textit{coarse-grained} and \textit{fine-grained} ones. Coarse-grained methods formulate the problem as a binary ~\cite{shi2015fast}\cite{wang2009unstructured} or triple classification~\cite{gaobiaoiv}, or road detection by labeling each pixel as road or non-road. They usually rely on prior rules, such as vanishing points~\cite{kong2009vanishing} or certain road models~\cite{alon2006off}.
However, the mechanical properties of autonomous vehicles are various, requiring a fine-grained understanding of terrain properties that can lead to a measure for the difficulty of terrain negotiation~\cite{wellhausen2019should}.
With the development of deep learning techniques in recent years, many deep semantic segmentation models are developed~\cite{minaee2021image}.
These models can be learned end-to-end on large-scale datasets with pixel-wise annotation, while both the size and diversity of the datasets are crucial to the model's performance~\cite{datahungry}.
Most of the open datasets in this scope describe urban scenes, such as Cityscapes~\cite{cordts2016cityscapes} and SemanticKITTI~\cite{behley2019semantickitti}. The few off-road ones ~\cite{valada16iser}\cite{wigness2019rugd} are of limited size and different definitions of semantic categories.
Fine-grained semantic segmentation in off-road scenes faces the following challenges: 1) There has been no unified category definition in nature scenes due to the diverse objects and ambiguous semantic interpretability; 2) Pixel-wise annotation of fine-grained labels is very hard because a large part of the pixels could suffer from severe semantic ambiguity, which makes manual annotation almost impractical; 3) Off-road scenes can be very different, and even in the same location, semantic properties can be changed greatly due to precipitations, temperature, defoliation, etc.

\begin{figure}[]
	\centering
	\includegraphics[width=0.46\textwidth]{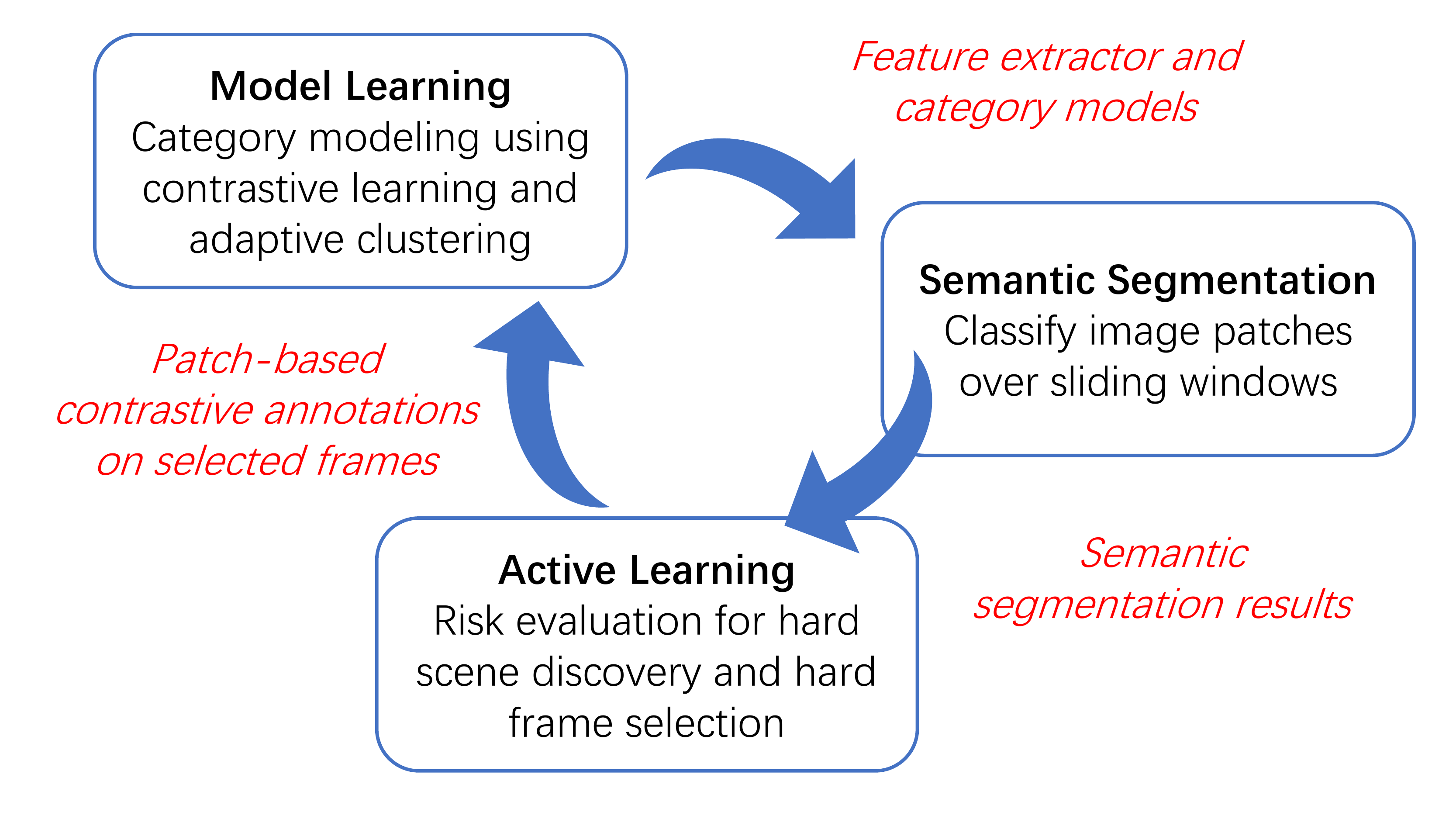}
	\caption{The proposed active and contrastive learning framework for fine-grained off-road semantic segmentation.}
	\label{fig:simple_pipeline}
\end{figure}
\begin{figure*}[t]
	\centering
	\includegraphics[width=\textwidth]{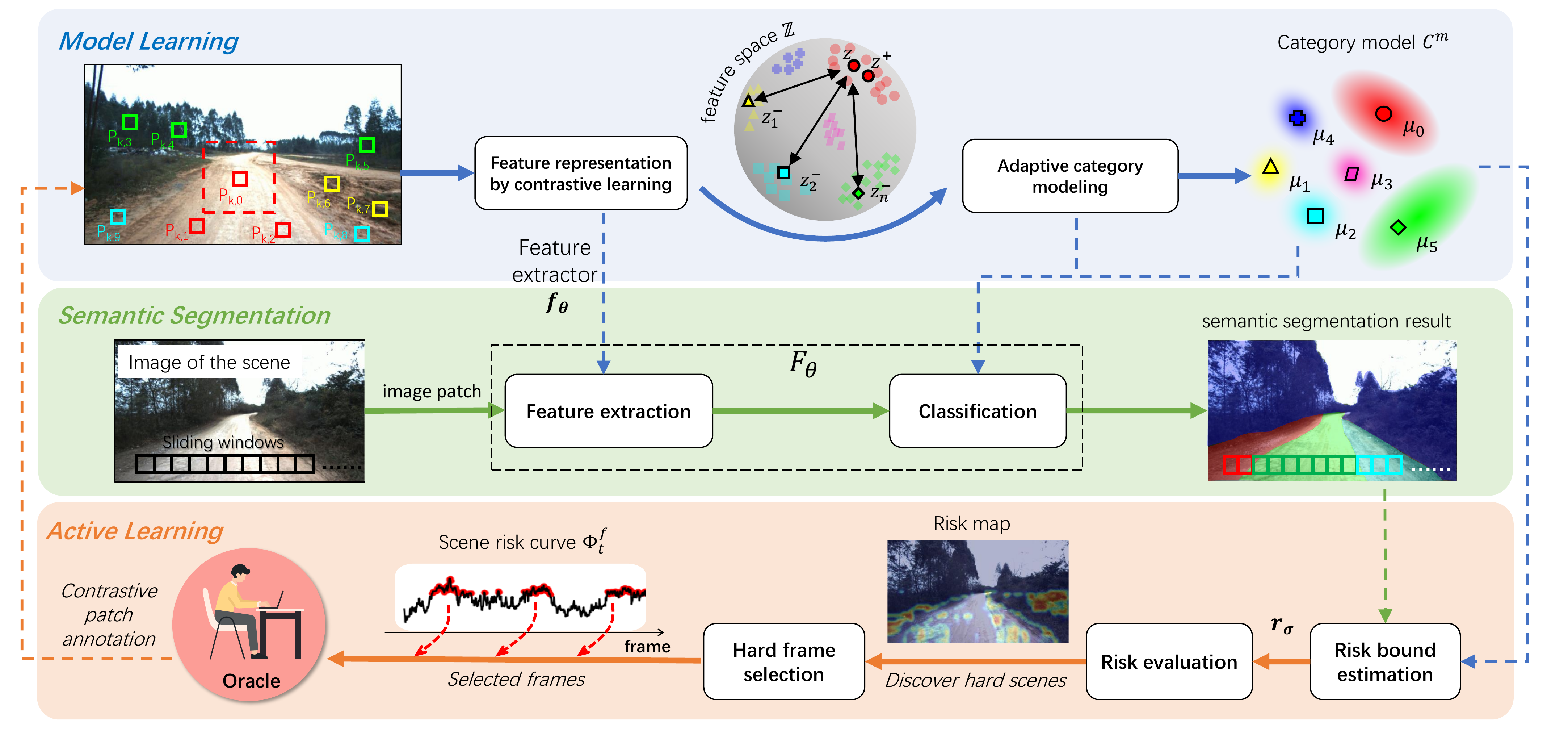}
	\caption{The proposed active and contrastive learning framework for fine-grained semantic segmentation of off-road scenes.}
	\label{fig:pipeline}
\end{figure*}

Facing the challenges, this research proposes a framework of fine-grained off-road semantic segmentation based on active and contrastive learning as illustrated in Fig.~\ref{fig:simple_pipeline} and detailed in Fig.~\ref{fig:pipeline}. 
It has the following features:
\begin{itemize}
	\item No pixel-wise annotated datasets: a patch-based annotation is devised to generate contrastive pairs of image patches that have different semantic attributes for weak supervision, and subsequently a sliding-window-based semantic segmentation is exploited;
	\item No predefined semantic categories: semantic categories are discovered and modeled on scene data by using contrastive learning for feature representation and adaptive clustering for category modeling;
	\item Adaptation to new scenes actively: a risk evaluation method is developed to discover scenes where the model results suffer from high-risk and the hard frames where the model is the most uncertain, so as to update the model actively and efficiently.
\end{itemize}
An off-road dataset is developed in this research containing three subsets of different scenes with a total of 8000 image frames. Extensive experiments are conducted to examine the performance of both key modules and the system flow of passive-active learning on both the self-developed and DeepScene~\cite{valada16iser} datasets. Experimental results show that a model of fine-grained off-road semantic segmentation can be learned through weak supervision on dozens of annotated image frames, when performance degradation is detected, active learning can be automatically triggered to update the model with additional annotations on no more than 40 selected hard frames. DeepScene experiments show that the proposed weakly supervised method achieves almost the same level of performance as the typical fully-supervised ones.

This paper is organized as follows. Related works are introduced in Section.~\ref{sec:2}. Section.~\ref{sec:3} describes the proposed contrastive and active learning method. In Section.~\ref{sec:4}, the experimental design are illustrated. Section.~\ref{sec:5} shows experimental results. Finally, we provide the conclusion in Section.~\ref{sec:6}.

\section{Related Works} \label{sec:2}
\subsection{Off-Road Semantic Segmentation}
Early researches were mainly coarse-grained, which are usually formulated as a binary classification problem. These methods depend on priors like vanishing point~\cite{kong2009vanishing}, vehicle trajectories~\cite{mei2017scene} or assume the road area as geometric shapes~\cite{zhou2010self}\cite{jeong2002vision}, or utilize fixed road models~\cite{alon2006off}\cite{wang2009unstructured}.

Benefiting from advances in deep learning, stronger feature representation results has led to fine-grained semantic segmentation capabilities. Rothrock et al.~\cite{rothrock2016spoc} firstly implemented FCN~\cite{long2015fully} for terrain classification of the Martian surface. After that, more studies~\cite{sgibnev2020deep,jin2021memory,viswanath2021offseg,guan2021ganav} focused on off-road scenes have been developed. Some of them deal with the challenges from various illumination and visual features in off-road scenes by combining multi-modal information with RGB images, such as stereo camera~\cite{chiodini2020evaluation}, NIR~\cite{valada16iser} and LiDAR~\cite{Maturana17}\cite{kim2018season}.
Due to the limitation of public datasets and the difficulty of off-road labeling, several studies tried to reduce the demand for fine-annotated data by transfer learning~\cite{holder2016road}\cite{sharma2019semantic} from urban or synthetic data.
For autonomous platforms with multiple sensors, weak supervision can be obtained from other modalities, such as LiDAR~\cite{tang2017one}\cite{gaobiaoiv}, audio features~\cite{zurn2020self} and force-torque signals~\cite{wellhausen2019should}. However, these automatically-generated labels are usually limited to certain specific categories and cannot meet fine-grained requirements.
In addition, few studies consider how to effectively adapt the model to the new off-road environments while avoiding pixel-wise annotations and network architecture changes.

\subsection{Contrastive Learning}
Contrastive learning has proven its promising ability to learn discriminative feature representations through a self-supervised pipeline by comparing positive and negative samples.
This idea has been widely used in many fields such as natural language processing~\cite{oord2018CPC} and typical visual tasks~\cite{tian2019contrastive}. These methods usually treat each instance and its augmented version as a positive pair, while other randomly selected instances are regarded as negative samples. In this setting, a large number of negative samples are required to ensure the effectiveness of the learned feature representation. The memory bank is usually used to store the features of the training data~\cite{tian2019contrastive,he2020momentum,wu2018unsupervised}.

Recent studies~\cite{NEURIPS2020_d89a66c7} proposed a supervised contrastive learning framework for the image classification task, which uses class labels to generate positive and negative samples. This idea has been extended to pixel-level semantic segmentation tasks by \cite{zhao2020contrastive}\cite{wang2021exploring}. However, pixel-wise annotations are extremely rare and expensive in off-road scenes.
Different from the setting of \cite{zhao2020contrastive}\cite{wang2021exploring}, this paper does not require pixel-level labels but only use a few sparse image patch-based annotations to distinguish similar or different regions of an image, and the features learned by contrastive learning are further used to generate fine-grained semantic segmentation results.

\subsection{Active Learning}
The core idea of active learning~\cite{settles2009active} is to let the trained model actively select the hardest or most informative samples to query manual annotations.
According to \cite{ren2021survey}, researches can be categorized by different query strategies: uncertainty-based approach~\cite{settles2008analysis,hwa2004sample,wang2016cost}, diversity-based approach~\cite{gal2017deep,nipsGuo10,nguyen2004active}, and expected model change~\cite{freytag2014selecting,roy2001toward,settles2007multiple}.
The uncertainty-based methods select samples with the highest uncertainty, which can be estimated by entropy~\cite{settles2008analysis}\cite{hwa2004sample} or softmax probability from deep neural networks~\cite{wang2016cost}. MC Dropout~\cite{gal2017deep} and ensemble methods~\cite{beluch2018power} can be used to improve uncertainty estimation.
Diversity-based methods~\cite{gal2017deep,nipsGuo10,nguyen2004active} tend to select samples in accord with input distributions, but it may lead to increased labeling costs.
The methods of expected model change~\cite{freytag2014selecting,roy2001toward,settles2007multiple} predict the influence of an unlabeled example on future model decisions, and choose the examples leading to more expected model change as an informative sample.

The active learning approaches for semantic segmentation usually use regions or entire images as the sampling unit.
Region-level methods~\cite{siddiqui2020viewal}\cite{mackowiak2018cereals} rely on a pre-segmentation to retrieve super-pixels, but due to insufficient or over-segmentation, the segmentation algorithm may not be able to separate appropriate semantic regions for labeling.
Image-level methods~\cite{yang2017suggestive}\cite{gorriz2017cost} use the entire image as the sampling unit. \cite{xie2020deal} incorporates the semantic difficulty to measure the informativeness and select samples at the image level. 
In this work, we also select samples at the image level. However, we do not require high-cost pixel-wise labels, but only query patch-wise weak annotations.

\section{Methodology}	\label{sec:3}

\subsection{Outline}
\begin{figure*}[t]
	\centering
	\includegraphics[width=0.8\textwidth]{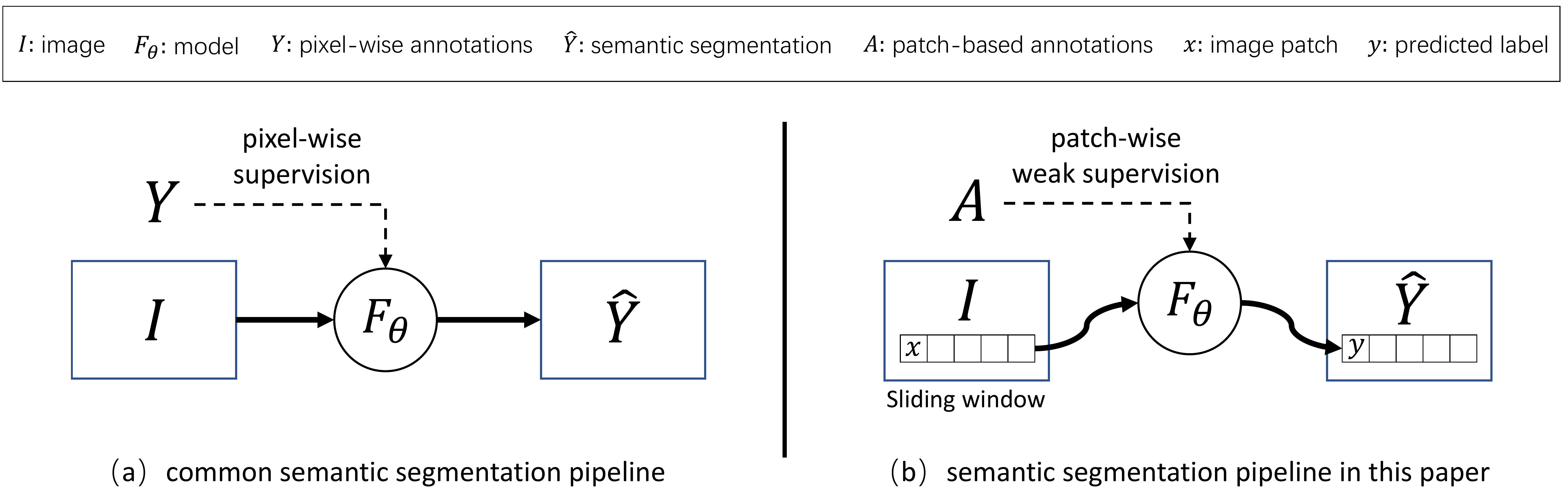}
	\caption{Semantic segmentation pipeline in this paper.}
	\label{fig:semantic_segmentation_def}
\end{figure*}

As illustrated in Fig.~\ref{fig:semantic_segmentation_def}(a), many deep semantic segmentation models $F_{\theta}$ take the entire image $I$ as input and map to semantic masks corresponding to each image pixels, which usually requires pixel-wise supervision. This research exploits a different flow as Fig.~\ref{fig:semantic_segmentation_def}(b) to take patch-based annotations as weak supervision. 
Given an image $I$, generate a sliding window and find the semantic label $y$ for each image patch $x$ through the classifier $F_{\theta}$.
In this research, $F_{\theta}$ consists of a feature extractor $f_{\theta}$ that discriminates a given contrastive image patches in the feature space, and a maximum likelihood classifier based on the category model $C^m$ that is learned by adaptive clustering of the training features. 

The model $F_{\theta}$ is trained at scene $D^{train}$ by a set of patch-based annotations $\{A_k\}$. At a new scene $D^{new}$, $F_{\theta}$ could be exposed to data that is substantially different from those in training, resulting in performance degradation. Such situation is very dangerous for safety-critical applications like autonomous driving. The agent needs to be aware of this performance degradation and require the model to be updated to accommodate the new scenes.
To this end, a risk evaluation method is developed to discover when the model is no longer valid and the results are high-risk, and triggers the process of active learning. In the active learning process, the hard frames which are the most uncertain for the model are tend to be selected for human annotation, and update the model $F_{\theta}$ subsequently.

\subsection{Model Learning}
\subsubsection{Problem Formulation}
One training image frame $I_k$ includes several anchor patch annotations $A_k=\{A_{k,i}=<p_{k,i},a_{k,i}>\}$. An anchor patch $A_{k,i}$ consists of an image patch $p_{k,i}$ and a label $a_{k,i}$. Different from common defined semantic labels that map a label ID to a specific category among the whole dataset, in this research, the labels of anchor patches are comparable only if they belong to the same image.
In other words, this $a_{k,i}$ only identifies image patches with similar or different semantic attributes in the current image. 
It provides great convenience for off-road data labeling, because it is difficult to determine a unified category list in advance for diverse off-road scenes.

Denoting $z=f_{\theta}(p)$ as an encoder that converts a high-dimensional image patch $p$ to a normalized $D$-dimensional feature vector $z\in \mathbb{Z}^D$, then the exponential cosine similarity $sim(p_i,p_j)$ is used to evaluate the similarity of two image patches via their features $z_i$ and $z_j$:

\begin{equation}\label{eqn_sim}
	sim(p_i,p_j)=exp(z_i^T \cdot z_j)
\end{equation}

The contrastive learning method is used to optimize $f_{\theta}$, which making the similarity between anchor patches $sim(p_{k,i},p_{k,j})$ be higher for $A_{k,i}$ and $A_{k,j}$ sharing the same label, i.e. $a_{k,i}=a_{k,j}$, and vice versa.

Through the optimized $f_{\theta}$ and extracted feature vectors $Z=\{z_1,z_2,...,z_N\}$, the category modeling aims to find the most applicable class number $m$ and corresponding model parameters $C^m=\{c_1,c_2,...,c_m\}$ by adaptive clustering.

\subsubsection{Feature Representation by Contrastive Learning} \label{sec:contrastive_learning}

\paragraph{Contrastive Samples}
The core idea behind contrastive learning is to learn a $f_\theta$ that separates samples with different semantic meanings.
At each step in the training process, contrastive learning requests a query sample $q$ following one corresponding positive sample $q^+$ and $n$ negative samples $\{q^-_i|i=1,..,n\}$.
Here, one query sample is an anchor patch $A_{k,i}$. Its corresponding positive and negative samples are all from the same image.

Given a query sample with label $a_{k,i}$ in image frame $I_k$, other anchor patches in $I_k$ can be divided to two sets according to the patch label $a_{k,i}$: one is positive anchor set $\{A_{k,i}^+\}$ with patches sharing the same label $a_{k,i}$; the other is negative anchor set $\{A_{k,i}^-\}$ including the rest patches. Positive and negative samples are selected from the aforesaid two sets respectively. The detailed sampling strategy is described in Section~\ref{sce:sampling}.

\paragraph{Network Design and Loss Function} \label{sec:nd_lf}

Use a convolutional neural network backbone to model $f_\theta$, i.e. AlexNet~\cite{krizhevsky2012imagenet}, convert the tensor of query, positive or negative samples into a normalized feature vector $z$ in low-dimensional embedding space $\mathbb{Z}^D$.
The parameter $\theta$ is optimized by contrastive learning, aiming to increase the exponential cosine similarity of $z$s that share the same label, while decreasing those with different labels.

A contrastive loss function InfoNCE~\cite{oord2018representation} is implemented:

\begin{equation}\label{eqn_loss}
	\mathcal{L}=-\log {\dfrac{\exp (z^T \cdot z^+/\tau)}{\exp (z^T \cdot z^+/\tau)+\sum_{i=1}^{n}{\exp (z^T \cdot z_i^-/\tau)}}}
\end{equation}
where $\tau$ denotes a temperature hyper-parameter, $z^+$ and $z_i^-$ are feature vectors of positive and negative samples.

Unlike the typical contrastive learning setting~\cite{Wu_2018_CVPR}, which uses a memory bank to save features of training samples, in this research, updated positive and negative sample features are calculated at each training step. This is because positive and negative samples are only comparable within the same image frame, which makes it possible to compute features with reasonable memory cost.

\paragraph{Sampling Strategy} \label{sce:sampling}

\begin{figure}[]
	\centering
	\includegraphics[width=0.5\textwidth]{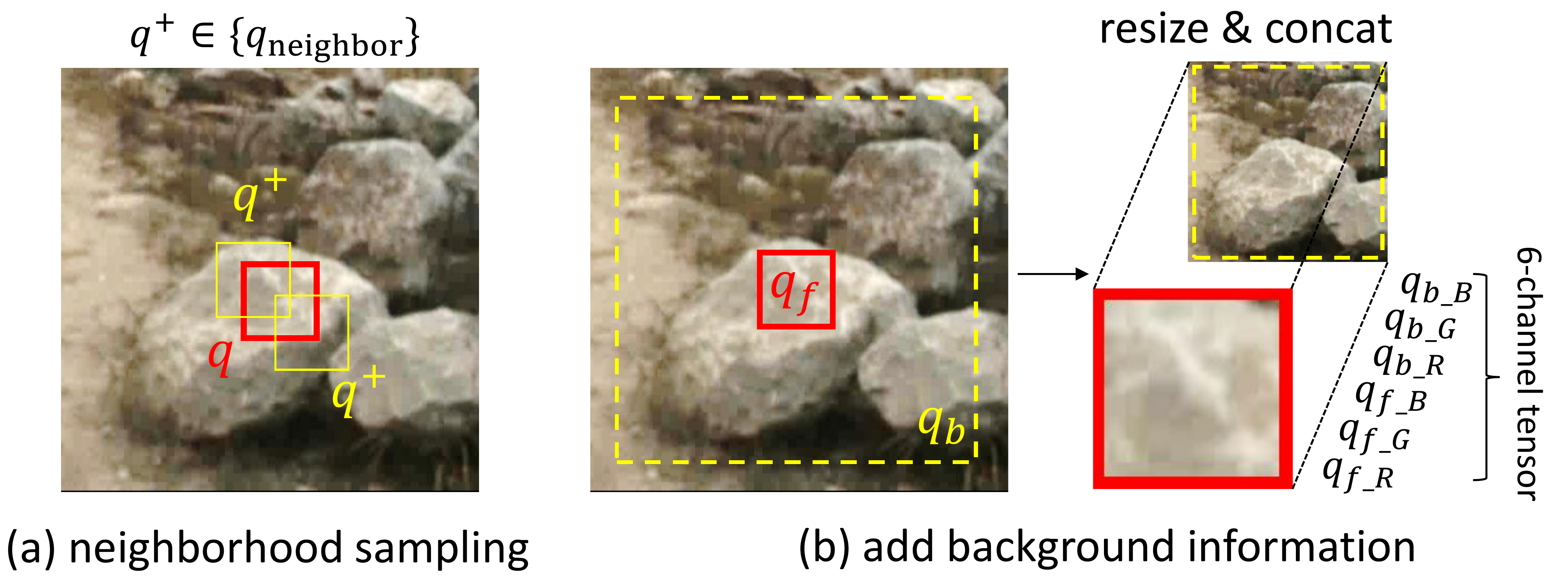}
	\caption{Illustration of (a) neighborhood sampling strategy, and (b) how to add background information with the foreground image patch.}
	\label{fig:dataaug}
\end{figure}

To enrich data variety of a limited number of anchor patches, assuming that neighbor regions in the off-road environment are semantically similar, positive and negative samples of a query sample $q$ are randomly drawn from the neighbor regions of $q$’s positive and negative anchors at the frame. As illustrated in Fig.~\ref{fig:dataaug}(a), the neighbor sample are drawn with its center point locating inside the region of the original sample $q$.

When composing a sample data, contextual information is also included, which is crucial for classifying objects that lack texture.
As shown in Fig.~\ref{fig:dataaug}(b), given an RGB image patch $q$, treating it as the foreground $q_f$ containing 3 channels, a background $q_b$ with a larger cropped area is centered at $q_f$. They are both reshaped to the same size and compose a 6-channel tensor as the input to the feature extractor $f_\theta$.
To enhance the robustness of the model in different environments with different illumination conditions, we implement data augmentation on the 6-channel tensor for each sample before feeding it to $f_\theta$. Concretely, data augmentation contains random greyscale, random flip and color jitter (randomly changing the brightness, contrast, and saturation of an image).

\subsubsection{Adaptive Category Modeling} \label{sec:category_modeling}
Given a set of $N$ $D$-dimensional data points $Z=\{z_1,z_2,...z_N\}$ that are feature vectors extracted by $f_{\theta}$ on image patches $\mathcal{P}=\{p_1,p_2,...,p_N\}$, category modeling is to find the category number $m$ and model parameters of the clusters $C^m=\{c_1,c_2,...,c_m\}$ that has the maximum likelihood on $Z$.

We consider the case where data following the multivariate Gaussian distribution, each cluster $c_k$ is modeled by a mean vector $\mu_k$ and a covariance matrix $\Sigma_k$. The likelihood of data point $z_i$ under cluster $c_k$ is

\begin{equation}
	\label{eqn_2}
	\gamma(z_i;c_k) = \displaystyle \frac{1}{(2\pi)^{D/2}|\Sigma_k|^{1/2}} \exp\{-\frac{1}{2}(z_i-\mu_k)^T \Sigma_k^{-1} (z_i-\mu_k)\}
\end{equation}

In the feature space $\mathbb{Z}^D$, the maximum likelihood function of the mixed category distribution $C^m$ on data points $Z$ can be given as follows~\cite{fraley1998algorithms},

\begin{equation}
	\label{eqn_3}
	L(C^m;Z) = \displaystyle \prod_{i=1}^{N} \sum_{k=1}^{m} {\gamma(z_i;c_k)},
\end{equation}.

For a certain cluster number $m$, the model parameters $C^m$ can be estimated by maximizing $L(C^m;Z)$, where the EM algorithm~\cite{mclachlan2019finite} is among the most popular approaches for parameter estimation.

To determine the number of clusters $m$, the Bayesian information criterion (BIC)~\cite{schwarz1978estimating} is used. Perform EM for each number of clusters $m=2,...,M$, where $M$ is an empirical value representing the maximal number of clusters. The BIC value is estimated as follows

\begin{equation}
	\label{eqn_4}
	BIC(C^m;Z) = -2\log L(C^m;Z)+u\log(N)
\end{equation}

where $u$ is the number of model parameters. The $m$ that leads to the decisive first local minimum of BIC value is found as the optimal number of clusters.

\subsection{Semantic Segmentation and Active Learning}

\subsubsection{Semantic Segmentation}
During semantic segmentation inference, for a given image frame $I_t$, a sliding window is conducted to generate image patches $\mathcal{P}_t=\{p_1,p_2,...p_{N_t}\}$. For each image patch $p_i$, a feature vector is first extracted by $z_i=f_{\theta}(p_i)$, then a category label $y_i$ is assigned by matching $z_i$ with the category model $C^m$ as below.

\begin{eqnarray}
	\label{eqn_5}
	k^* &=& \displaystyle \arg\max_k (\gamma(z_i;c_k)) \\
	\label{eqn_6}
	r_i &=& 1-\gamma(z_i;c_{k^*}) \\
	\label{eqn_7}
	y_i &=&
	\left\{
	\begin{array}{ll}
		k^* &if \ r_i \leq \mathbf{r_{\sigma}}; \\
		\phi &otherwise \\
	\end{array}
	\right.
\end{eqnarray}

where $k^*$ is the cluster ID that has the maximal likelihood with $z_i$. $r_i$ is the risk of classifying $z_i$ to the most likely cluster $k^*$. $k^*$ is assigned to category label $y_i$ if and only if the risk $r_i$ is below a certain risk bound $\mathbf{r_{\sigma}}$. Otherwise, $y_i$ will be assigned a special label $\phi$ indicating the unknown class. 

It is generally believed that all pixels in one patch belong to the same category. When acquiring image patches by a $s_p \times s_p$ sliding window with step size $\xi$, if higher resolution semantic segmentation is required, usually set $\xi < s_p$, so that the patches are partially overlapped. As a result, each pixel may get multiple predictions from different patches. The final label of each pixel is determined by the weighted voting method. When calculating, the closer a pixel is to the center of the patch, the higher the voting weight of the patch label.

After obtaining the semantic segmentation of the entire image, the widely used Dense Conditional Random Field (DenseCRF~\cite{krahenbuhl2011efficient}) is used as an optional post-processing module to refine the predictions.
In scenes with clear region boundaries, segmentation can be effectively refined.

\subsubsection{Risk Bound Estimation}
After learning the category model $C^m$ on the training set $\mathcal{D}^{train}$ including a total of $N^{train}$ patches set $\mathcal{P}^{train}$, a confidence level $\delta$ of the learned model can be given by the operator's experience. 
It means that the proportion of risky classification is less than $1-\delta$. Therefore, the risk bound $\mathbf{r_{\sigma}}$ can be estimated by solving the following constrained optimization.

\begin{equation}
	\label{eqn_8}
	\left\{
	\begin{array}{ll}
	\min \  &\mathbf{r_{\sigma}} \\
	\textit{s.t.} & \frac {|\{r_i > \mathbf{r_{\sigma}}\}|}{N^{train}} \leq 1-\delta
	\end{array}
	\right.
\end{equation}

\begin{figure}[]
	\centering
	\includegraphics[width=0.35\textwidth]{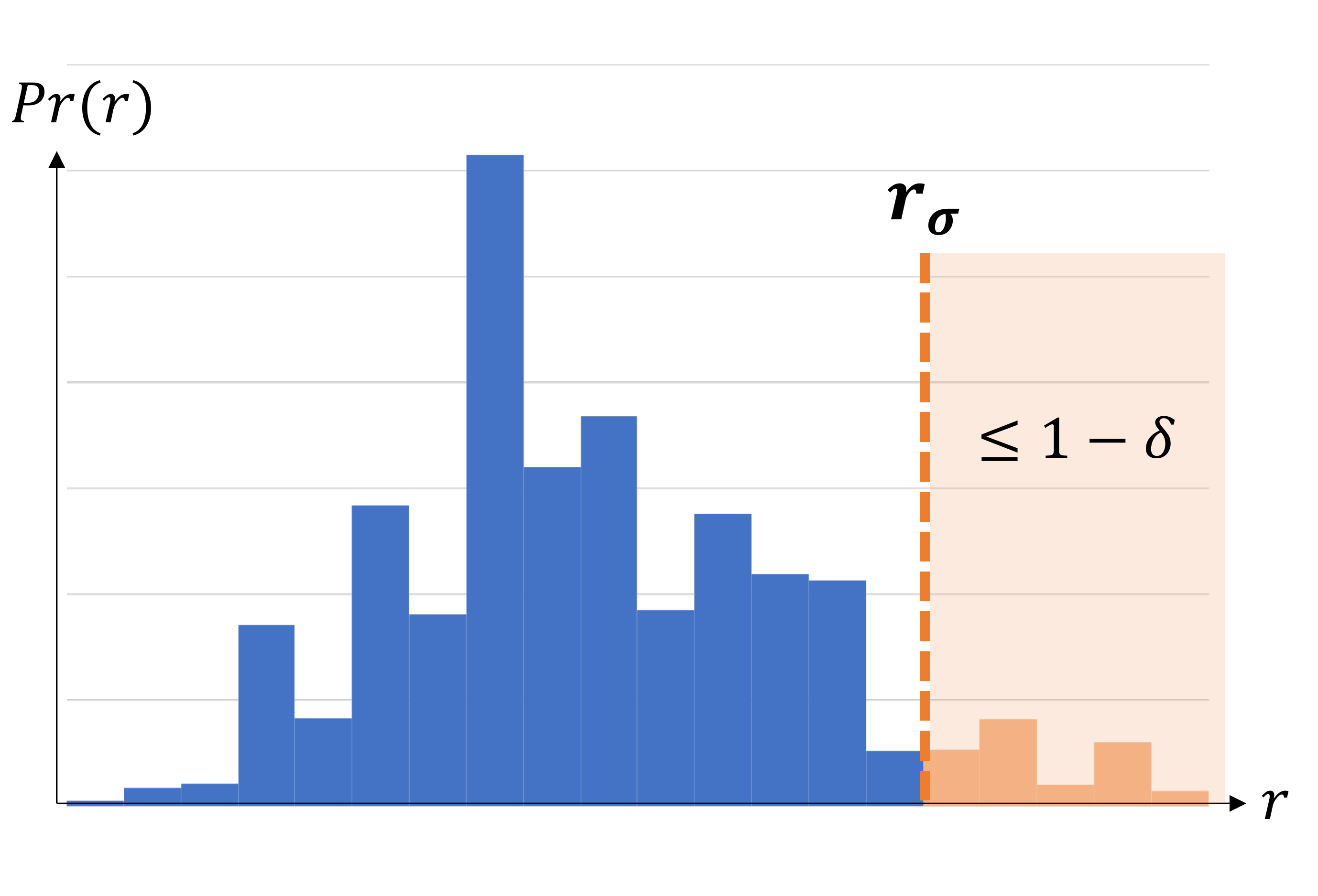}
	\caption{Illustration of risk bound estimation.}
	\label{fig:histogram}
\end{figure}

An analytical solution is shown in Fig.~\ref{fig:histogram}.
Generate a histogram $Pr(r)$ over the set of $\{r_i\}$, where each $r_i$ is computed over $p_i \in \mathcal{P}^{train}$. Minimize $\mathbf{r_{\sigma}}$ while satisfying $\frac {|\{r_i > \mathbf{r_{\sigma}}\}|}{N^{train}} \leq 1-\delta$, equivalent to $\sum_{r>r_{\sigma}}^{r_{\max}}Pr(r) \leq 1-\delta$. The resulting risk bound $\mathbf{r_{\sigma}}$ is used for the following risk evaluation.

\subsubsection{Risk Evaluation}

During inference, given a set of image patches $\mathcal{P}_t=\{p_1,p_2,...,p_{N_t}\}$ belonging to image frame $I_t$, along with the category model $C^m$ with $m$ clusters and a risk bound $\mathbf{r_{\sigma}}$, a set of labels $Y_t=\{y_1,y_2,...y_{N_t}\}$ will be estimated, where $y_i \in \{1,...,m;\phi\}$. For convenience, we denote the classification of each data point by $y_i=\mathbf{y}(p_i|C^m)$.
Let $\mathcal{P}_t^{*} \subset \mathcal{P}_t$ be the subset containing patches classified as $\phi$,

\begin{equation}
	\label{eqn_9}
	\mathcal{P}_t^{*} = \mathcal{P}_t^{*}(\mathcal{P}_t;C^m) = \{p_i \in \mathcal{P}_t \wedge \mathbf{y}(p_i|C^m)=\phi\}
\end{equation}

The proportion of $\mathcal{P}_t^{*}$ represents the degree of uncertainty or risk of the model in the scene described by the image frame $I_t$. An index describing the model uncertainty at the image frame level, i.e.\textbf{ frame-level risk}, is subsequently defined:
\begin{equation}
	\label{eqn_10}
	\Phi_t^f = \Phi_t^f(\mathcal{P}_t;C^m) = \displaystyle \frac {|\mathcal{P}_t^{*}(\mathcal{P}_t;C^m)|}{N_t},
\end{equation}
where $|\mathcal{P}_t^{*}(\mathcal{P}_t;C^m)|$ is the size of the set $\mathcal{P}_t^{*}$.

When deploying on the autonomous vehicle, the data to be predicted is usually a sequence of $T_s$ image frames $\{I_1,...,I_{T_s}\}$ including image patches $S=\{\mathcal{P}_1,...,\mathcal{P}_{T_s}\}$. Let $S^{*} \subset S$ be the subset containing risky frames that the model is uncertain, i.e. exceeding the risk level $\epsilon$.

\begin{equation}
	\label{eqn_11}
	S^{*} = S^{*}(S;C^m) = \{P_t \in S \wedge \Phi_t^f(P_t;C^m)>\epsilon\}
\end{equation}

The proportion of $S^{*}$ represents the degree of risk, in other words, the uncertainty of the model on the dataset described by the sequence of image frames $S$. An index describing \textbf{sequence-level risk} is then defined.

\begin{equation}
	\label{eqn_12}
	\Phi^s = \Phi^s(S;C^m) = \displaystyle \frac {|S^{*}(S;C^m)|}{T_s}
\end{equation}

In this research, sequence-level risk is a measure of discovering when a model is no longer valid and trigger the active learning process, while frame-level risk is to find the hard frames that the model is uncertain, which are requested for human annotation.

\subsubsection{Workflow} \label{sec:hard_scene_selection}

The workflow of active learning for semantic segmentation is described below.

\begin{itemize}
	\item [$\mathbb{S}$1. ]\textit{Offline learning}
	\begin{itemize}
		\item [$\mathbb{S}$1-1. ]\textit{Initialization (Model learning)}
		
		Given a training dataset including anchor patches set $A^{train}$, learn a model $f_\theta$ by contrastive learning (Section~\ref{sec:contrastive_learning}), then find category model $C^m$ by adaptive clustering (Section~\ref{sec:category_modeling}).
		
		\item [$\mathbb{S}$1-2.] \textit{Risk Bound Estimation}
		
		Given the learned category model $C^m$ on the training patches  $\mathcal{P}^{train}$ from $A^{train}$, estimate a risk bound $\mathbf{r}_\sigma$ (Eqn.~\ref{eqn_8}).
	\end{itemize}

	\item [$\mathbb{S}$2. ] \textit{Online Semantic Segmentation}
	
	\begin{itemize}
	\item [$\mathbb{S}$2-1. ] \textit{Semantic Segmentation}
	
	Given a test image frame $I_t$, generate a set of image patches $\mathcal{P}_t$ using a sliding window. For each image patch $p_i$, find the corresponding label $y_i$ (Eqn.~\ref{eqn_7}) and risk $r_i$ (Eqn.~\ref{eqn_6}).
	
	\item [$\mathbb{S}$2-2.] \textit{Risk Evaluation}
	
	For consecutive $T_s$ test image frames $\{I_1,...,I_{T_s}\}$ including image patches $S=\{\mathcal{P}_t\}$, estimate $\{\mathcal{P}^{*}_t\}$ (Eqn.~\ref{eqn_9}), $\{\Phi^f_t\}$ (Eqn.~\ref{eqn_10}), $S^{*}$ (Eqn.~\ref{eqn_11}) and $\Phi^s$ (Eqn.~\ref{eqn_12}). If the sequence-level risk $\Phi^s$ exceeds a certain threshold, the active leaning module will be triggered to update the current model.
	\end{itemize}
	
	\item [$\mathbb{S}$3. ] \textit{Active Learning}
	\begin{itemize}
		
%
	
	\item [$\mathbb{S}$3-1.] \textit{Hard Frame Selection}
	
	Choose a batch of $\mathcal{B}$ image frames $\{I_{u_1},...,I_{u_\mathcal{B}}\}$ with image patches $S_u=\{\mathcal{P}_{u_1},...,\mathcal{P}_{u_\mathcal{B}}\} \subset S$ to meet
	\begin{equation}
	\left\{
	\begin{array}{ll}
		\label{eqn_13}
	 	\max \ & \sum_{t=u_1}^{u_\mathcal{B}}{\Phi^f_t} \\
		\textit{s.t.} & \forall \mathcal{P}_{u_i}, \mathcal{P}_{u_j} \in S_u, |u_i-u_j|>\Delta
	\end{array}
	\right.,
	\end{equation}
	
	where $\Delta$ is a threshold to avoid selection on neighbor frames that provide repetitive information.  
	
	\item [$\mathbb{S}$3-2.] \textit{Human Annotation}
	
	For selected $\mathcal{B}$ frames, annotate contrastive image patches and corresponding labels to obtain $\mathcal{A} = \{A_{u_1},...,A_{u_\mathcal{B}}\}$, where $A_u=\{<p_{u,i}, a_{u,i}>\}$, then update the supplemental annotations $A^{AL} \gets A^{AL}+\mathcal{A}$.
	
	\item [$\mathbb{S}$3-3.] \textit{Model Update}
	
	Fine-tune $f_\theta$ on the $A^{AL}$, then estimate category model $C^m$ and risk bound $\mathbf{r}_{\sigma}$. Finally, continue semantic segmentation process (go to $\mathbb{S}$2-1).
	\end{itemize}
\end{itemize}

\section{Experimental Design} \label{sec:4}
\subsection{Notations}
\subsubsection{Experiment Stage}
For clear interpretation, the following notations are introduced:
\begin{itemize}
	\item \textit{Learning}: the initial training stage of semantic segmentation models. Given a training set containing anchor annotations, train the feature extractor $f_\theta$ and the corresponding category model $C^m$ by contrastive learning.
	\item \textit{Active Learning}: Given a learned model, implement it on a new dataset and use active learning to obtain supplemental annotations, then update the model.
	\item \textit{Test}: evaluate the performance of semantic segmentation models.
\end{itemize}
\subsubsection{Model}
The notations of semantic segmentation models:
\begin{itemize}
	\item $M_A$: the model trained on dataset $A$;
	\item $M_A^B$: based on the initial model $M_A$, the model updated by active learning on dataset $B$.
\end{itemize}
When emphasizing the semantic granularity of annotations, the following notations are used:
\begin{itemize}
	\item $M_{A^{Lv1}}$: the model trained by annotations with $Lv1$ (Level 1) semantic granularity on dataset $A$. In experiment, the granularity of anchor annotations include 3 levels, i.e. $Lv1$, $Lv2$ and $Lv3$.
\end{itemize}
When indicating the frame number for training, the following notations are used:
\begin{itemize}
	\item $M_{A50}$: the model trained by 50 image frames of dataset $A$;
	\item $M_{A50}^{B20}$: based on the initial model $M_A50$, the model updated by active learning on 20 frames of dataset $B$.
\end{itemize}
When comparing different methods, $M$ is usually replaced by the method's abbreviation, such as $Base_A$, $Rand_{A50}^{B20}$ and $Ours_{A50}^{B20}$.

\subsection{Dataset}

\begin{figure}[]
	\centering
	\includegraphics[width=0.48\textwidth]{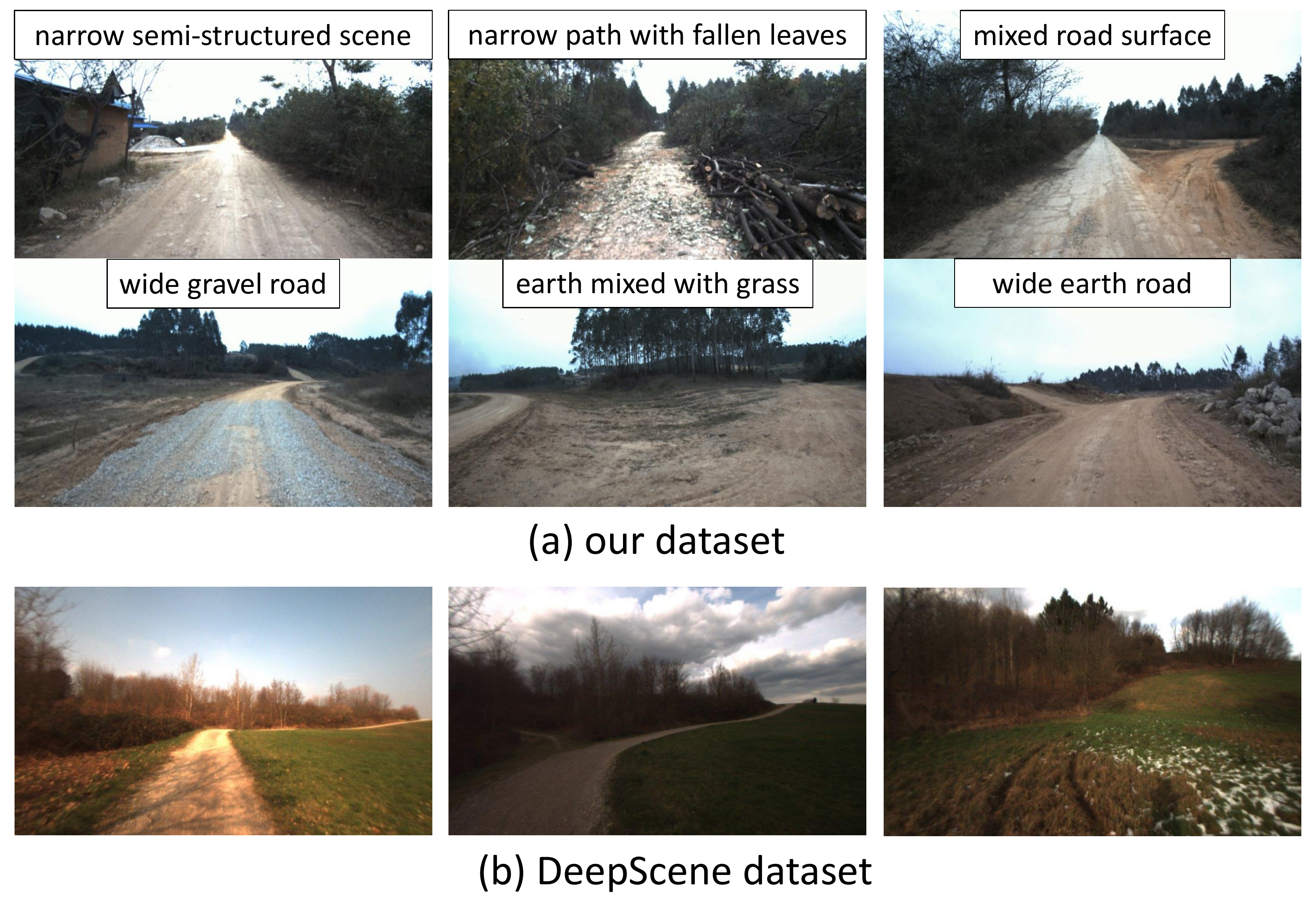}
	\caption{Typical scenes in (a) our off-road dataset and (b) DeepScene dataset~\cite{valada16iser}.}
	\label{fig:dataset}
\end{figure}

\subsubsection{Our Dataset}
We developed an off-road dataset for experimental validation of the proposed method. As shown in Fig.~\ref{fig:dataset}(a), the images are collected by a front-view monocular RGB camera mounted on a moving vehicle. 
As shown in Table~\ref{tab:dataset}, our dataset includes 3 sub-datasets (noted as A/B/C) for different experimental stages. 

\textit{Subset A} contains 5064 frames for experiments of the \textit{Learning} stage. We randomly sample 50 frames for patch-based annotations and use them to train a contrastive learning-based feature extractor and obtain adaptive category modeling, evaluating its performance by randomly selecting patches on other images.

\textit{Subset B} includes 1639 frames for evaluating the active learning pipeline. The risk evaluation module will check the results of $M_A$. When the active learning is activated, $X$ frames will be selected from hard frames for human annotation and the model is updated to $M_A^B$.

\textit{Subset C} includes 1600 frames for evaluating the improvements from active learning. Concretely, the effectiveness of active learning can be examined by comparing the semantic segmentation results and risk-based metrics between the initial model $M_A$ and the updated model $M_A^B$.

\begin{table}[]
	\caption{Dataset statistics}
	\label{tab:dataset}
	\centering
	\renewcommand{\arraystretch}{1.2}
	\begin{threeparttable}
		\begin{tabular}{c|ccc|cc}
			\hline
			\textbf{Dataset}                                                     & \multicolumn{3}{c|}{\textbf{Ours}}                                                                                          & \multicolumn{2}{c}{\textbf{DeepScene}}                           \\ \hline
			\begin{tabular}[c]{@{}c@{}}Subset code\end{tabular}                & A                                                        & B                                                         & C    & D                                                         & E    \\ \hline
			\begin{tabular}[c]{@{}c@{}}Exp. stage\end{tabular}           & Learning & \begin{tabular}[c]{@{}c@{}}Active\\ learning\end{tabular} & Test & \begin{tabular}[c]{@{}c@{}}Active\\ learning\end{tabular} & Test \\ \hline
			\begin{tabular}[c]{@{}c@{}}Annotated \\ anchor frames\end{tabular} & 50                                                       & 10$\sim$30                                                & -    & 20$\sim$40                                                & -    \\ \hline
			\begin{tabular}[c]{@{}c@{}}Total frames\end{tabular}               & 5064                                                     & 1639                                                      & 1600 & 230                                                       & 136  \\ \hline
		\end{tabular}
		\begin{tablenotes}
			\footnotesize
			\item[*] The train and test set of DeepScene dataset are noted as D and E respectively.
		\end{tablenotes}
	\end{threeparttable}
\end{table}

\begin{figure}[]
	\centering
	\includegraphics[width=0.5\textwidth]{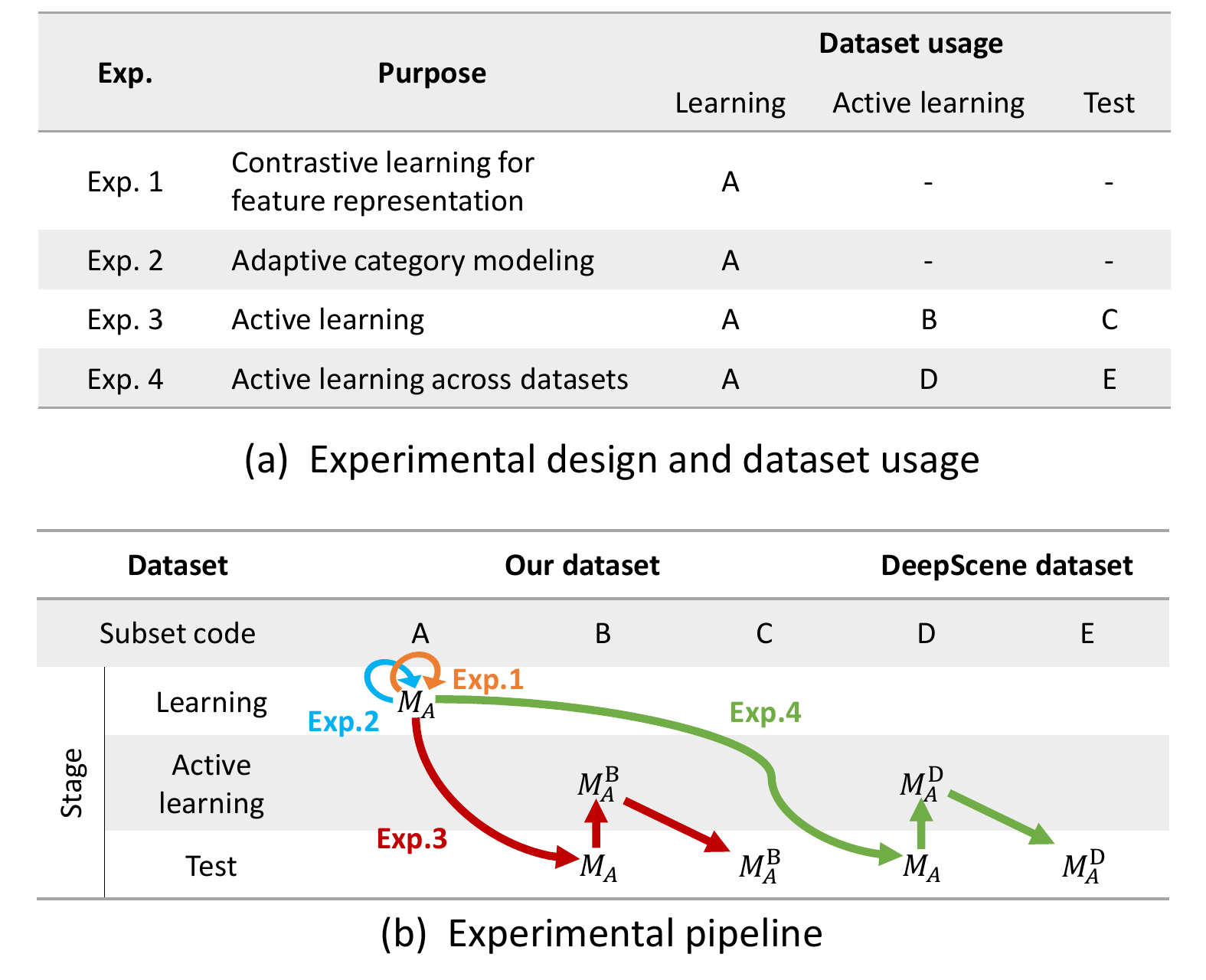}
	\caption{Experimental design and pipeline. (a) experimental pipeline. (b) experimental design and dataset usage.}
	\label{fig:exp_pipeline}
\end{figure}

\subsubsection{DeepScene Dataset}
Besides our dataset, the proposed method is also evaluated on the DeepScene dataset~\cite{valada16iser}. It contains several types of camera data, while this paper only uses monocular RGB images. As shown in Fig.~\ref{fig:dataset}(b), compared to our dataset,the DeepScene dataset has different environments and illumination to examine the generalization ability of the proposed method in different scenes.

The train and test set of DeepScene are noted as dataset D and E, including 230 and 136 frames respectively. Each image frame has pixel-wise semantic annotations. Similar to dataset B/C, the dataset D/E are respectively used for training and testing of the active learning models.

\subsection{Experimental Design}

Four experiments are designed as shown in Fig.~\ref{fig:exp_pipeline}, which are introduced as follows:

\subsubsection{Exp.1 Contrastive learning for Feature Representation}
It aims to evaluate the proposed contrastive learning method for feature extraction. Train the model $M_A$ by annotations on 50 randomly selected frames from dataset A. The results are shown in Section~\ref{sec:exp1}.

\subsubsection{Exp.2 Adaptive Category Modeling}
It is designed to examine the results of adaptive category modeling. In the experiments, we use dataset A labeled by 3 granularity levels, noted from coarse-grained to fine-grained as $A^{Lv1}$, $A^{Lv2}$ and $A^{Lv3}$. The results are shown in Section~\ref{sec:exp2}.

\subsubsection{Exp.3 Active Learning}
It aims to evaluate the proposed active learning pipeline. The initial model $M_A$ trained on dataset A is deployed on dataset B. After activating the active learning module to select a few image frames for human annotation, the model is updated from $M_A$ to $M_A^C$ and tested on dataset C. The results are shown in Section~\ref{sec:exp3}.

\subsubsection{Exp.4 Active Learning across Datasets}
It aims to demonstrate the cross-dataset generalization ability of the proposed active learning method. The initial model $M_A$ is deployed on dataset D. After activating the active learning module to select a few image frames for human annotation, the model is updated from $M_A$ to $M_A^D$ and tested on dataset E. The results are shown in Section~\ref{sec:exp4}.

\begin{figure*}[]
	\centering
	\includegraphics[width=0.75\textwidth]{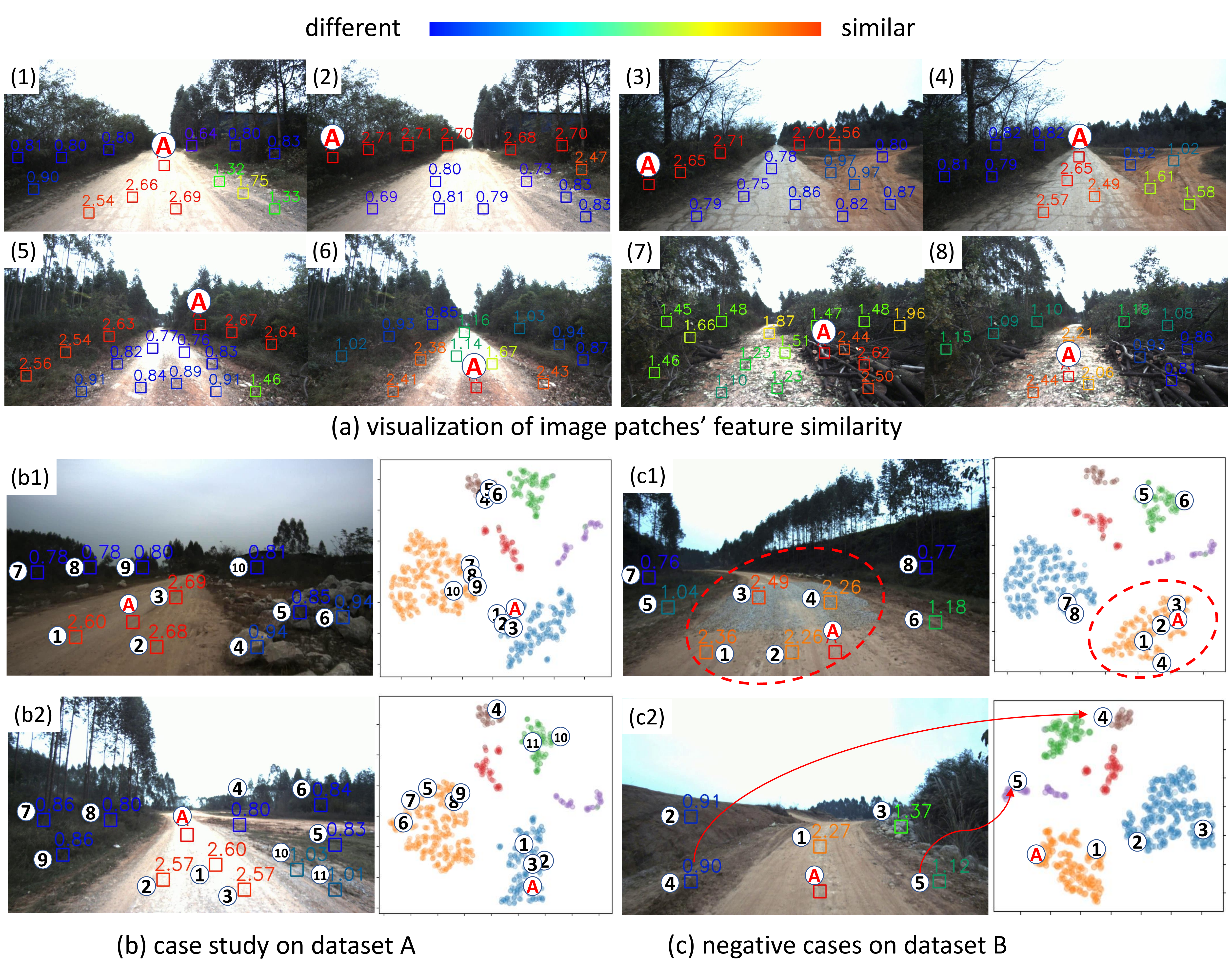}
	\caption{Visualization of image patches' feature similarity. Randomly choose one patch as an anchor patch (red A), other patches' color indicate their similarity to the anchor patch (Equation~\ref{eqn_sim}). (a) image patches' feature similarity at different scenes. (b) case study on dataset A: the corresponding positions of the image patches in the feature space. (c) negative cases on dataset B: feature similarity does not match semantic meanings.}
	\label{fig:exp1}
\end{figure*}

\subsection{Evaluation Metrics}
On our dataset, two risk-based metrics are used to evaluate models' performance.
\begin{itemize}
	\item \textbf{FLR} (frame-level risk). Note the frame-level risk $\Phi_t^f$ (Equation~\ref{eqn_10}) as \textbf{FLR}, then \textbf{mFLR} indicates the mean \textbf{FLR} of a data sequence, which describes the average proportion of high risk patches per frame. The lower \textbf{FLR} value means the better model performance.
	
	\item \textbf{Sc} (scene coverage), defined as \textbf{Sc}$=1-\Phi_t^s$, where sequence-level risk $\Phi_t^s$ is described by Equation~\ref{eqn_12}. It means the certainty of the model over the entire data sequence, i.e. the proportion of non-risk frames. The higher \textbf{Sc} value means the better performance.
\end{itemize}

On the DeepScene dataset, 5 common metrics for semantic segmentation tasks are introduced:
\begin{itemize}
\item \textbf{mIoU} (mean Intersection over Union), \textbf{PA} (Pixel Accuracy), \textbf{PRE} (Precision), \textbf{REC} (Recall), \textbf{FPR} (False Positive Rate).
\end{itemize}
\begin{figure}[]
	\centering
	\includegraphics[width=0.45\textwidth]{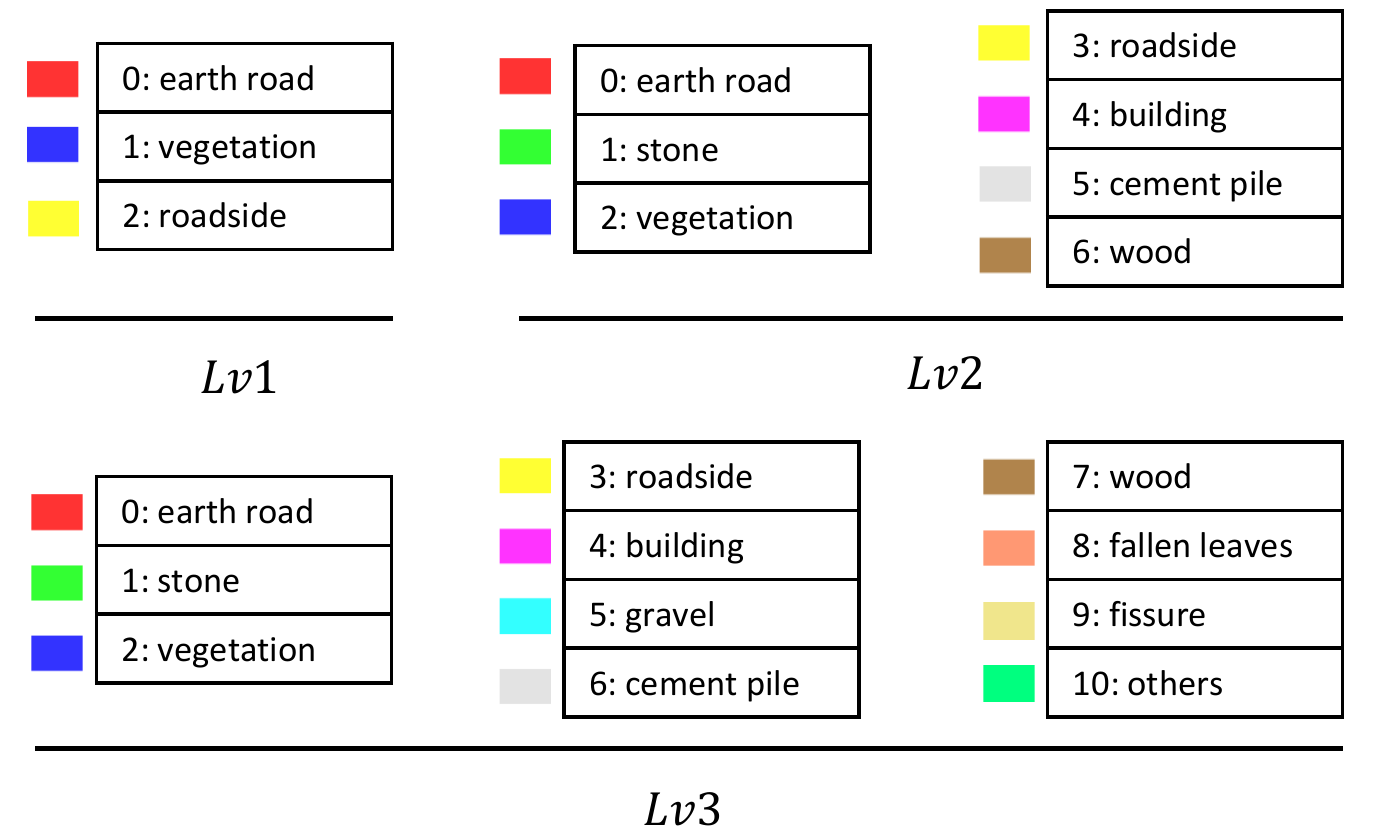}
	\caption{The reference label definitions of different semantic granularity.}
	\label{fig:exp2_label_define}
\end{figure}
\begin{figure}[]
	\centering
	\includegraphics[width=0.5\textwidth]{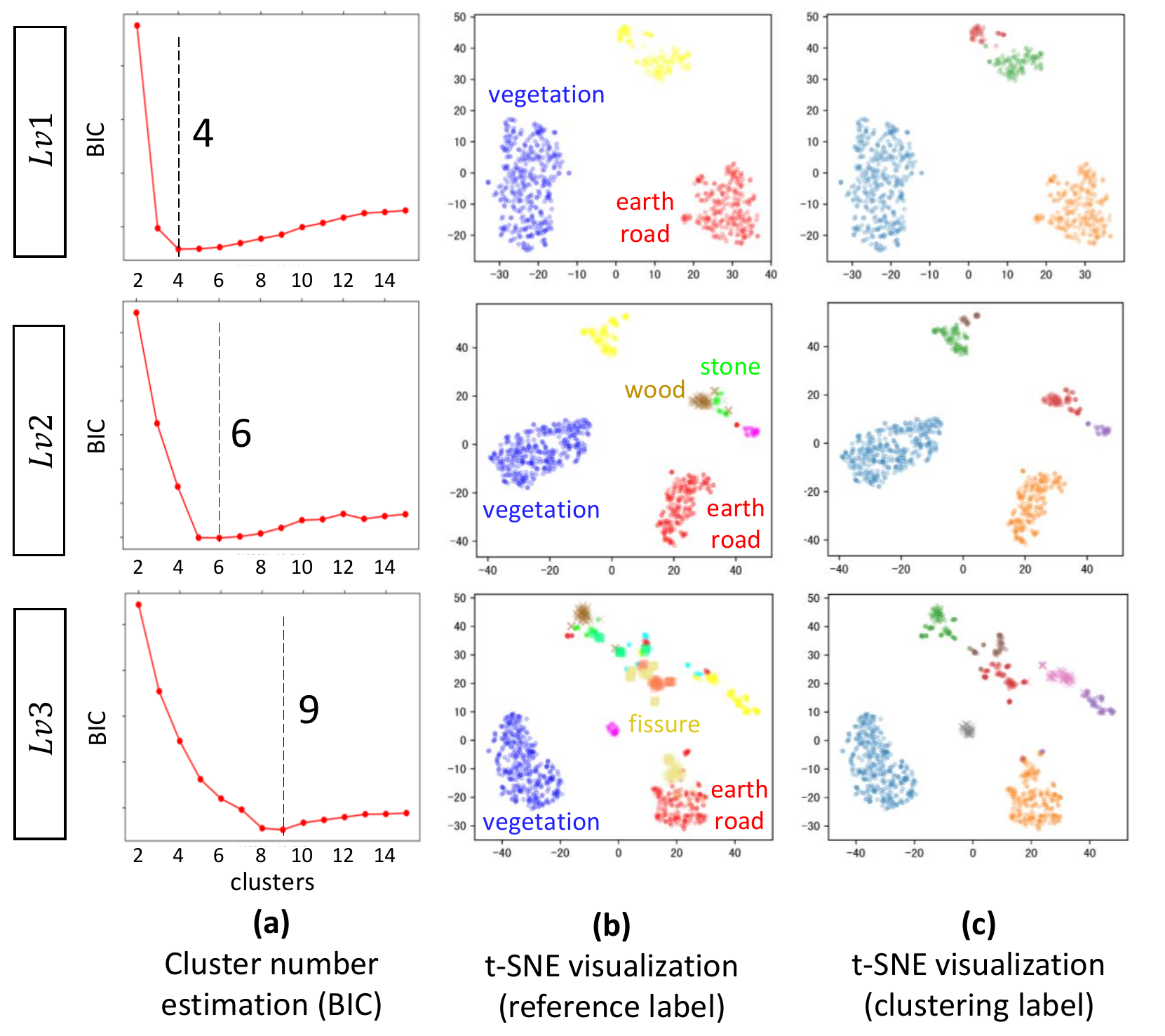}
	\caption{The cluster number estimated by BIC and adaptive category modeling results under different granularity annotations. (a) cluster number estimation. (b) category modeling results colorized by reference labels. (c) category modeling results colorized by clustering labels.}
	\label{fig:exp2_tsne}
\end{figure}

\begin{figure*}[]
	\centering
	\includegraphics[width=0.75\textwidth]{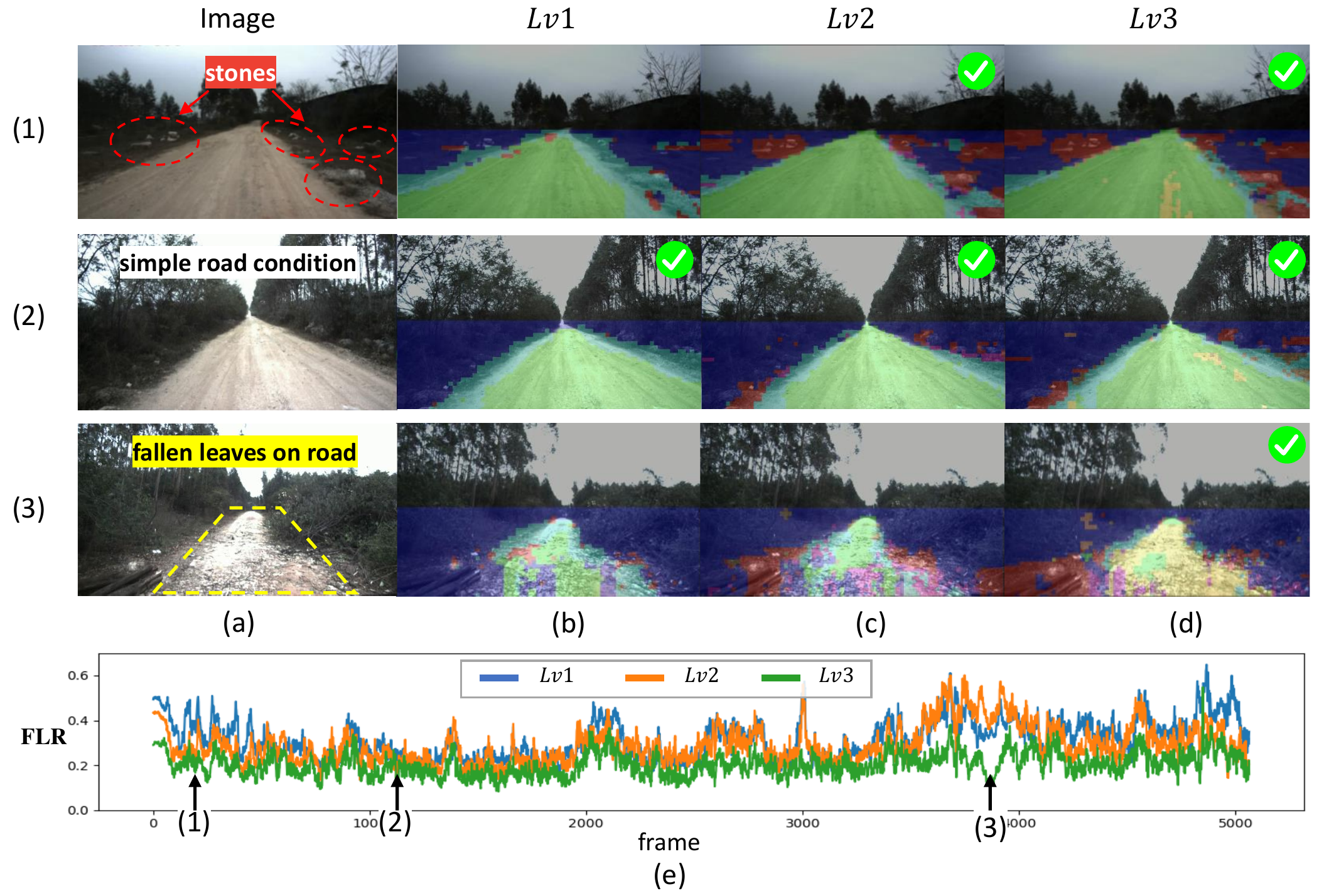}
	\caption{Analysis of adaptive category modeling results under different granularities. (a) input image; (b-d) predictions from the model under granularity $Lv1$-$Lv3$; (e) The \textbf{FLR} curves of models trained by three granularities on dataset A.}
	\label{fig:exp2_example}
\end{figure*}

\begin{figure}[]
	\centering
	\includegraphics[width=0.5\textwidth]{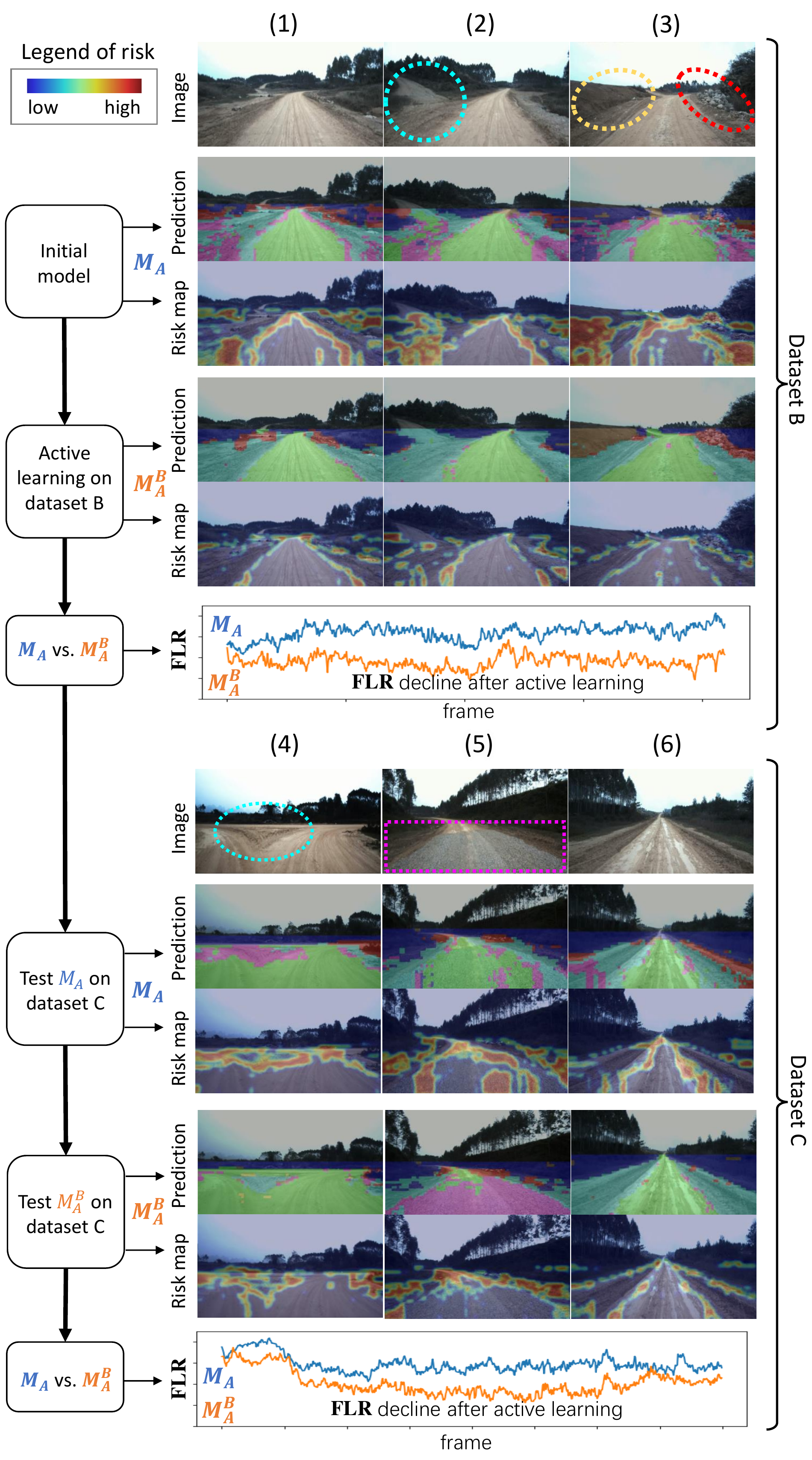}
	\caption{Exp.~3: Comparison of model's semantic segmentation results before and after active learning.}
	\label{fig:exp3}
\end{figure}

\begin{figure*}[]
	\centering
	\includegraphics[width=\textwidth]{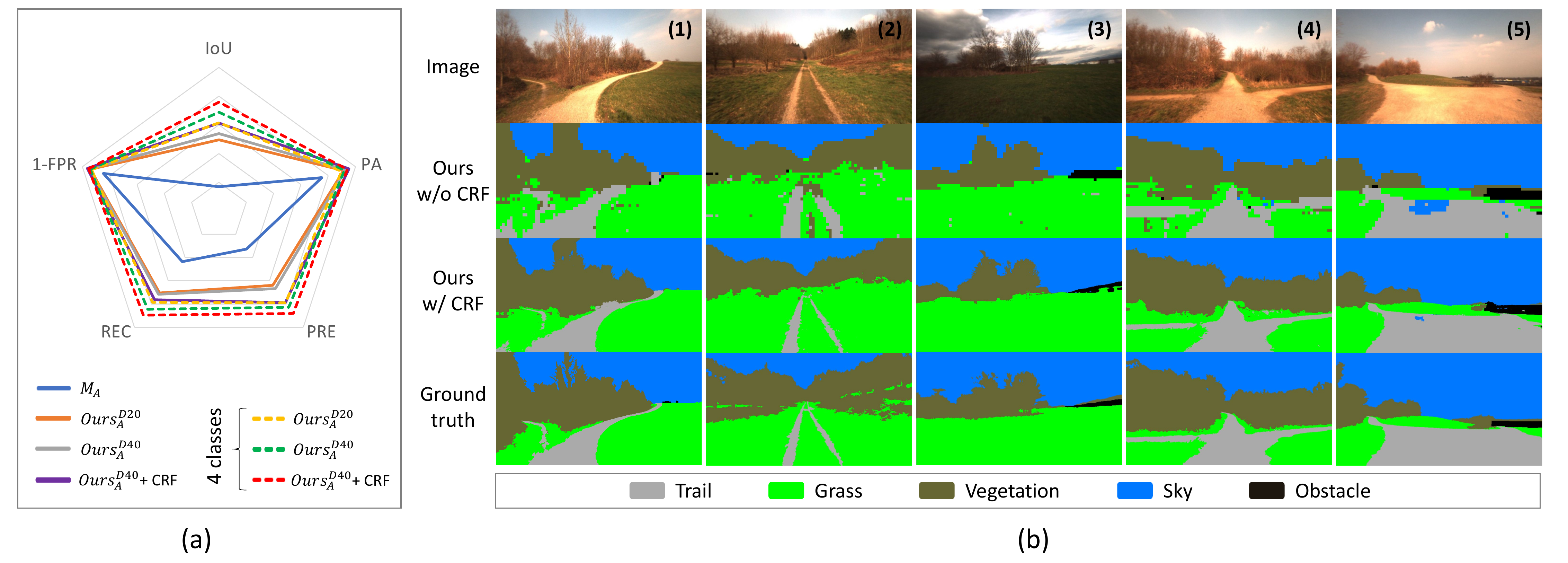}
	\caption{Semantic segmentation results on the DeepScene dataset. (a) the radar chart visualization of methods' performance shown in Table~\ref{tab:DeepScene}. (b) qualitative semantic segmentation results of the proposed method.}
	\label{fig:exp4}
\end{figure*}

\section{Experimental Results} \label{sec:5}

\subsection{Exp.~1: Contrastive Learning for Feature Representation} \label{sec:exp1}
In the initial training stage of the semantic segmentation model, the feature extractor $f_\theta$ trained by contrastive learning is dedicated to narrowing down the semantically similar image patches in the feature space, while pushing  away different image patches.
Fig.~\ref{fig:exp1} shows the patch similarity (Equation~\ref{eqn_sim}) calculated by the feature extractor of model $M_A$ on non-training image frames. Similarity values are marked above the image patches.

As shown in Fig.~\ref{fig:exp1}(a), several image patches are randomly selected in each image, one of which is regarded as an anchor patch (marked as red A). Different colors visualize feature similarities between the anchor patch and other patches. Colors closer to red indicate higher similarity.

Fig.~\ref{fig:exp1}(b) selects two cases in dataset A for concrete analysis. The feature vector of each image patch is reduced by t-SNE\parencite{van2008visualizing}, and then drawn on the right. The patch IDs in the two images correspond to each other. The other points in the feature map are from the training patches.
For both cases in Fig.~\ref{fig:exp1}(b), the feature similarity between each image patch and anchor is consistent with their semantic relationship, and image patches of the same category are relatively concentrated in the feature map. Such distribution provides a prerequisite for subsequent adaptive category modeling.

Fig.~\ref{fig:exp1}(c) shows some negative cases of model $M_A$ in an unknown scenario (dataset B). In Fig.~\ref{fig:exp1}(c1), the gravel road (patch ID 3,4) in the red circle does not appear in the training set, while the earth road samples (patch ID 1,2) have a high-level similarity with them and distribute close in the feature map. It indicates that the current features extracted by $f_\theta$ are difficult to distinguish between the two categories.
In Fig.~\ref{fig:exp1}(c2), the image patch 4 and 5 (muddy areas on the roadside) of the same category are distributed farther in the feature map, which also reflects the weakness of $f_\theta$. In the new scenes, the generalization ability of the features extractor is insufficient, and active learning needs to be introduced to update the model.

\subsection{Exp.~2: Adaptive Category Modeling} \label{sec:exp2}

The mechanical properties of unmanned platforms are different, which requests different granularities for semantic segmentation, i.e. the types of terrain that need to be distinguished are also different.
According to the mechanical properties of the platform, we define three reference label sets with different granularities, as shown in Fig.~\ref{fig:exp2_label_define}.
It needs to be emphasized that it is very difficult to annotate pixel-wise labels based on these label sets due to the ambiguity between classes. However, they can provide guidance for labeling patch-based positive and negative samples.

Fig.~\ref{fig:exp2_tsne}(a) shows the result of using Bayesian Information Criterion (BIC) to determine the number of clusters in adaptive category modeling.
Under the three granularity annotations ${Lv1}$, ${Lv2}$ and ${Lv3}$, the training samples are adaptively clustered into 4, 6, and 9 categories, respectively.
Figure ~\ref{fig:exp2_tsne} (b-c) is the visualization result of the image patch features in the training data after dimensional reduction by t-SNE\parencite{van2008visualizing}, which is colored according to the reference label and clustering label.

From the clustering distribution of Fig.~\ref{fig:exp2_tsne}(b-c), under all three semantic granularities, \textit{earth road}, \textit{vegetation} and other dominated categories adaptive clustering results are basically consistent with the reference label.
Small-sized clusters are usually from categories with low frequency, such as \textit{gravel}, \textit{fallen leaves}, etc.

There exist differences between these clustering results and the reference labels, which are mainly divided into two situations: one is the category with complex and diverse appearance. For example, the \textit{fissure} samples in Fig.~\ref{fig:exp2_tsne}(b)-$Lv3$ has multiple scattered clusters, indicating high variance within the category.
The second situation is the category with similar features to other clusters. For example, in Fig.~\ref{fig:exp2_tsne}(b)-$Lv2$, the distribution of \textit{wood} samples is relatively concentrated, but very close to \textit{stone} samples. In category modeling, they can easily be misclassified into the same category.

Concrete cases are shown in Fig.~\ref{fig:exp2_example}. The \textit{stones} in case (1) have been effectively modeled at granularity $Lv2$ and $Lv3$, but failed to be segmented at coarse-grained $Lv1$.
In case (2), such simple road condition only includes basic categories like earth roads and vegetation. The corresponding semantic segmentation results under different granularity are basically the same. From the \textbf{FLR} curve in Fig.~\ref{fig:exp2_example}(e), it can also be seen that the risk values of different models in case (2) are close and low, indicating that a basic category modeling is enough to handle such simple scenes.
In case (3), the roads covered by \textit{fallen leaves} are marked as yellow in the semantic segmentation results. Only the fine-grained $Lv3$ model can distinguish this category. From the corresponding \textbf{FLR} curves in Fig.~\ref{fig:exp2_example}(e), it can be seen that the curve of $Lv3$ is significantly reduced. The fine-grained category modeling effectively reduces prediction risk.
\begin{table}[]
	\caption{Performance of active learning methods}
	\label{tab:al_compare}
	\centering
	\renewcommand{\arraystretch}{1.8}
	\setlength\tabcolsep{5pt}
	\begin{threeparttable}
		\begin{tabular}{c|c|cc|cc}
			\hline
			&                  & \multicolumn{2}{c|}{\textbf{mFLR} $\downarrow$}                                & \multicolumn{2}{c}{\textbf{Sc}}                                                \\ \hline
			& model            & Dataset B                         & Dataset C                         & Dataset B                         & Dataset C                         \\ \hline
			\multirow{4}{*}{\rotatebox{90}{\makecell[c]{comparison of \\frame selection \\strategy}}} & $M_{A}$          & 54.25\%                           & 59.04\%                           & 35.50\%                           & 8.75\%                            \\
			& $Rand_{A}^{B20}$ & 46.52\%                           & 49.50\%                           & 78.00\%                           & 66.25\%                           \\
			& $Unif_{A}^{B20}$ & 44.43\%                           & 44.68\%                           & 83.63\%                           & 73.00\%                           \\
			& $Ours_{A}^{B20}$ & \textbf{44.34\%} & \textbf{40.67\%} & \textbf{85.00\%} & \textbf{83.88\%} \\ \hline
			\multirow{5}{*}{\rotatebox{90}{\makecell[c]{comparison of \\labeled frame \\numbers}}}    & $Ours_{A}^{B10}$ & 51.57\%                           & 50.70\%                           & 44.88\%                           & 61.50\%                           \\
			& $Ours_{A}^{B15}$ & 45.58\%                           & 40.58\%                           & 77.88\%                           & 81.50\%                           \\
			& $Ours_{A}^{B20}$ & 44.34\%                           & 40.67\%                           & 85.00\%                           & 83.88\%                           \\
			& $Ours_{A}^{B25}$ & 43.03\%                           & 38.12\%                           & 88.13\%                           & 84.75\%                           \\
			& $Ours_{A}^{B30}$ & \textbf{39.22\%} & \textbf{35.36\%} & \textbf{91.00\%} & \textbf{86.38\%} \\ \hline
		\end{tabular}
		\begin{tablenotes}
			\footnotesize
			\item[1] All models' subscript $A50$ is abbreviated to $A$;
			\item[2] $Rand_{A}^{B20}$: random select frames for active learning; $Unif_{A}^{B20}$: uniformly select frames for active learning;
			\item[$\downarrow$]: lower value means better performance.
		\end{tablenotes}
	\end{threeparttable}
\end{table}
\begin{table}[]
	\caption{Performance on DeepScene dataset}
	\label{tab:DeepScene}
	\centering
	\renewcommand{\arraystretch}{1.6}
	\setlength\tabcolsep{5pt}
	\begin{threeparttable}
		\begin{tabular}{c|c|lccccc}
			\hline
			type                               & class              & \multicolumn{1}{c}{model} & PA    & IoU   & PRE   & REC   & FPR$\downarrow$ \\ \hline
			\multirow{7}{*}{\rotatebox{90}{weakly sup.}} & \multirow{4}{*}{5} & $M_A$                     & 75.28 & 16.95 & 32.65 & 43.43 & 15.35           \\
			&                    & $Ours_{A}^{D20}$          & 91.30 & 49.60 & 63.84 & 70.40 & 5.92            \\
			&                    & $Ours_{A}^{D40}$          & 92.66 & 53.96 & 66.76 & 71.56 & 5.02            \\
			&                    & $Ours_{A}^{D40}$+CRF      & \textbf{95.16} & \textbf{61.26} & \textbf{78.72} & \textbf{76.30} & \textbf{3.67}            \\ \cline{2-8} 
			& \multirow{3}{*}{4} & $Ours_{A}^{D20}$          & 89.81 & 61.51 & 78.85 & 78.56 & 6.76            \\
			&                    & $Ours_{A}^{D40}$          & 91.89 & 68.81 & 82.69 & 84.42 & 5.73            \\
			&                    & $Ours_{A}^{D40}$+CRF      & \textbf{94.01} & \textbf{75.88} & \textbf{88.12} & \textbf{89.52} & \textbf{4.57}            \\ \hline
			\multirow{3}{*}{\rotatebox{90}{fully sup.}}                   &  \multirow{3}{*}{5}     & SegNet~\cite{badrinarayanan2017segnet}                   & 88.47 & 74.81 & 84.63 & 86.39 & 13.53           \\
			&                    & FCN~\cite{long2015fully}                       & 90.95 & 77.46 & 87.38 & 85.97 & 10.32           \\
			&                    & ParseNet~\cite{liu2015parsenet}                 & 93.43 & 83.65 & 90.07 & 91.57 & 8.94            \\ \hline
		\end{tabular}
		\begin{tablenotes}
			\footnotesize
			\item[1] \textbf{IoU}: Intersection over Union, \textbf{PA}: Pixel Accuracy, \textbf{PRE}: Precision, \textbf{REC}: Recall, \textbf{FPR}: False Positive Rate.
		\end{tablenotes}
	\end{threeparttable}
\end{table}

\subsection{Exp.~3: Results of Active Learning} \label{sec:exp3}

Fig.~\ref{fig:exp3} shows the semantic segmentation results at different stages in the active learning process. First, deploy the model $M_A$ originally trained on dataset A on dataset B, trigger the active learning module to select a small number of frames in difficult scenes, and after manual supplementary annotation, the model is updated to $M_A^B$, then compare its performance to $M_A$.
Finally, the semantic segmentation performance of the model $M_A$ and $M_A^B$ are compared on dataset C. Fig.~\ref{fig:exp3}(1-3) and (4-6) show the semantic segmentation results and \textbf{FLR} curves of the two models on dataset B and C respectively.

In Fig.~\ref{fig:exp3}(1-3), it can be found that the predictions of $M_A$ on dataset B contains many noises.
For example, the lime soil inside the cyan circle in Fig.~\ref{fig:exp3}(2) and the stones inside the red circle in Fig.~\ref{fig:exp3}(3) are predicted to be a mixture of multiple categories, and the corresponding positions in risk maps also present a high prediction risk (red). In Fig.~\ref{fig:exp3}(3), the earth embankment inside the yellow circle is a new type for model $M_A$, and its segmentation results are noisy and the risk value is also high.
After active learning, the model $M_A^B$ gives better semantic segmentation results in aforesaid scenes, and the high-risk areas in the risk map are also significantly reduced.
Among the dataset, the reduction of frame-level risk is reflected by the \textbf{FLR} curve. After active learning, the orange curve of $M_A^B$ is significantly lower than the blue curve of $M_A$.

On dataset C, we also test and compare the performance of model $M_A$ and $M_A^B$. The muddy triangle area in Fig.~\ref{fig:exp3}(4), the gravel road in Fig.~\ref{fig:exp3}(5), etc., all show the better semantic segmentation performance of $M_A^B$, which is also reflected in the lower \textbf{FLR} curve on dataset C.

Table~\ref{tab:al_compare} quantitatively compares the performance of different models. The upper part of the table compares different frame selection strategies. Each model selects 20 frames on dataset B for active learning. $Rand_A^{B20}$ randomly selects frames for annotation, $Unif_{A}^{B20}$ uses uniform sampling, and $Ours_{A}^{B20}$ uses the hard frame selection strategy proposed in Section~\ref{sec:hard_scene_selection}.
Under the metrics \textbf{mFLR} and \textbf{Sc}, the proposed method shows significant advantages.
The bottom half of Table~\ref{tab:al_compare} evaluates the effect of the active sampled frame number $\mathcal{B}$. In general, more supplemented annotations lead to better model performance.

\subsection{Exp.~4: Results of Active Learning across Datasets} \label{sec:exp4}

Table~\ref{tab:DeepScene} is the performance comparison of different models on DeepScene dataset. Based on the pixel-wise labels of DeepScene, several classical semantic segmentation metrics are evaluated for comparison.
According to the category number, models are divided into two groups:
(1) models with 5 categories, consistent with the label definitions of DeepScene dataset;
(2) models with 4 categories based on the adaptive category modeling, the ignored label corresponds to \textit{obstacle} in DeepScene.
\textit{Obstacle} samples are very rare in the training set of DeepScene, accounting for only $0.33\%$. The active learning module only selects a few image frames, which makes it difficult for \textit{obstacles} to get annotations. Therefore, it is not considered as a valid independent label during adaptive category modeling.

In the table, the 5-class model $Ours_A^{D20}$ uses only 20 frames of patch-based weak annotations in the new environment, while achieving 16.02\% PA and 32.65\% IoU improvement over the initial model $M_A$. The model $Ours_A^{D40}$ further improves the metrics.
Since the road boundaries in the DeepScene dataset are relatively clear, the post-processing module DenseCRF\parencite{krahenbuhl2011efficient} can refine the semantic segmentation results. The corresponding model $Ours_{A}^{D40}$+CRF achieves the overall best performance. Under both label definitions, the proposed active learning method brings significant performance gains.

Comparing models with different category numbers: when the numbers of frames for supplementary annotation are the same, the 4-class model obtained by adaptive category modeling outperforms the 5-class model in all indicators.
Among them, the optimal models under both category definitions have better FPR and PA than some classic fully supervised methods. 
The 4-class model $Ours_{A}^{D40}$+CRF achieves the same level as the fully supervised methods in all evaluation metrics.

Fig.~\ref{fig:exp4}(a) uses a radar chart to visualize model performance in Table~\ref{tab:DeepScene}. Some concrete examples are shown in Fig.~\ref{fig:exp4}(b). In various scenes under different lighting conditions, the model $Ours_{A}^{D40}$+CRF (5 categories) requires only 40 frames of low-cost patch-based annotations, while achieving good semantic segmentation results.

\section{Conclusion}   \label{sec:6}
In this paper, we propose a framework of fine-grained off-road semantic segmentation based on active and contrastive learning.
Through patch-based weak annotations, a contrastive learning-based feature extractor is learned to discriminate different semantic attributes. After that, an adaptive category modeling method is proposed, then a sliding-window-based semantic segmentation is exploited. To help the model adapt to new scenes efficiently, a risk evaluation method is developed to discover and select hard frames for active learning of new scenes.
To evaluate the proposed method, extensive experiments are conducted on the self-developed off-road dataset with a total of 8000 image frames and the public DeepScene dataset. With only dozens of image frames as weak supervision, the fine-grained off-road semantic segmentation model can be learned. 
When detecting performance degradation in new scenes, the proposed active learning method can effectively select hard frames for the current model by risk evaluation and improve results with no more than 40 frames of patch-based annotations. Experiments on the DeepScene dataset show that the proposed weakly supervised method can achieve the same level of performance as typical fully supervised ones.
Future work will be addressed on preventing models from catastrophic forgetting after adapting to new scenes. Research about continual learning and incremental learning will be explored in the future.


\ifCLASSOPTIONcaptionsoff
  \newpage
\fi

\printbibliography

@ARTICLE{datahungry,
	author={Gao, Biao and Pan, Yancheng and Li, Chengkun and Geng, Sibo and Zhao, Huijing},
	journal={IEEE Transactions on Intelligent Transportation Systems}, 
	title={Are We Hungry for 3D LiDAR Data for Semantic Segmentation? A Survey of Datasets and Methods}, 
	year={2021},
	pages={1-19}}

@inproceedings{siam2017deep,
	title={Deep semantic segmentation for automated driving: Taxonomy, roadmap and challenges},
	author={Siam, Mennatullah and Elkerdawy, Sara and Jagersand, Martin and Yogamani, Senthil},
	booktitle={International Conference on Intelligent Transportation Systems},
	pages={1--8},
	year={2017},
	organization={IEEE}
}

@article{settles2007multiple,
	title={Multiple-instance active learning},
	author={Settles, Burr and Craven, Mark and Ray, Soumya},
	journal={Advances in Neural Information Processing Systems},
	volume={20},
	pages={1289--1296},
	year={2007}
}

@article{roy2001toward,
	title={Toward optimal active learning through monte carlo estimation of error reduction},
	author={Roy, Nicholas and McCallum, Andrew},
	journal={International Conference on Machine Learning},
	volume={2},
	pages={441--448},
	year={2001}
}

@inproceedings{freytag2014selecting,
	title={Selecting influential examples: Active learning with expected model output changes},
	author={Freytag, Alexander and Rodner, Erik and Denzler, Joachim},
	booktitle={European Conference on Computer Vision},
	pages={562--577},
	year={2014},
	organization={Springer}
}

@inproceedings{nguyen2004active,
	title={Active learning using pre-clustering},
	author={Nguyen, Hieu T and Smeulders, Arnold},
	booktitle={International Conference on Machine Learning},
	pages={79},
	year={2004}
}

@inproceedings{nipsGuo10,
	author    = {Yuhong Guo},
	title     = {Active Instance Sampling via Matrix Partition},
	booktitle = {Advances in Neural Information Processing Systems},
	pages     = {802--810},
	publisher = {Curran Associates, Inc.},
	year      = {2010}
}

@article{mackowiak2018cereals,
	title={Cereals-cost-effective region-based active learning for semantic segmentation},
	author={Mackowiak, Radek and Lenz, Philip and Ghori, Omair and Diego, Ferran and Lange, Oliver and Rother, Carsten},
	journal={arXiv preprint arXiv:1810.09726},
	year={2018}
}

@inproceedings{siddiqui2020viewal,
	title={Viewal: Active learning with viewpoint entropy for semantic segmentation},
	author={Siddiqui, Yawar and Valentin, Julien and Nie{\ss}ner, Matthias},
	booktitle={IEEE Conference on Computer Vision and Pattern Recognition},
	pages={9433--9443},
	year={2020}
}

@article{gorriz2017cost,
	title={Cost-effective active learning for melanoma segmentation},
	author={Gorriz, Marc and Carlier, Axel and Faure, Emmanuel and Giro-i-Nieto, Xavier},
	journal={arXiv preprint arXiv:1711.09168},
	year={2017}
}

@inproceedings{xie2020deal,
	title={Deal: Difficulty-aware active learning for semantic segmentation},
	author={Xie, Shuai and Feng, Zunlei and Chen, Ying and Sun, Songtao and Ma, Chao and Song, Mingli},
	booktitle={Asian Conference on Computer Vision},
	year={2020}
}

@inproceedings{yang2017suggestive,
	title={Suggestive annotation: A deep active learning framework for biomedical image segmentation},
	author={Yang, Lin and Zhang, Yizhe and Chen, Jianxu and Zhang, Siyuan and Chen, Danny Z},
	booktitle={International conference on medical image computing and computer-assisted intervention},
	pages={399--407},
	year={2017},
	organization={Springer}
}

@inproceedings{beluch2018power,
	title={The power of ensembles for active learning in image classification},
	author={Beluch, William H and Genewein, Tim and N{\"u}rnberger, Andreas and K{\"o}hler, Jan M},
	booktitle={IEEE Conference on Computer Vision and Pattern Recognition},
	pages={9368--9377},
	year={2018}
}

@inproceedings{gal2017deep,
	title={Deep bayesian active learning with image data},
	author={Gal, Yarin and Islam, Riashat and Ghahramani, Zoubin},
	booktitle={International Conference on Machine Learning},
	pages={1183--1192},
	year={2017},
	organization={PMLR}
}

@article{wang2016cost,
	title={Cost-effective active learning for deep image classification},
	author={Wang, Keze and Zhang, Dongyu and Li, Ya and Zhang, Ruimao and Lin, Liang},
	journal={IEEE Transactions on Circuits and Systems for Video Technology},
	volume={27},
	number={12},
	pages={2591--2600},
	year={2016},
	publisher={IEEE}
}

@article{hwa2004sample,
	title={Sample selection for statistical parsing},
	author={Hwa, Rebecca},
	journal={Computational linguistics},
	volume={30},
	number={3},
	pages={253--276},
	year={2004},
	publisher={MIT Press One Rogers Street, Cambridge, MA 02142-1209, USA}
}

@inproceedings{settles2008analysis,
	title={An analysis of active learning strategies for sequence labeling tasks},
	author={Settles, Burr and Craven, Mark},
	booktitle={Conference on Empirical Methods in Natural Language Processing},
	pages={1070--1079},
	year={2008}
}

@article{ren2021survey,
	title={A survey of deep active learning},
	author={Ren, Pengzhen and Xiao, Yun and Chang, Xiaojun and Huang, Po-Yao and Li, Zhihui and Gupta, Brij B and Chen, Xiaojiang and Wang, Xin},
	journal={ACM Computing Surveys (CSUR)},
	volume={54},
	number={9},
	pages={1--40},
	year={2021},
	publisher={ACM New York, NY}
}

@inproceedings{wu2018unsupervised,
	title={Unsupervised feature learning via non-parametric instance discrimination},
	author={Wu, Zhirong and Xiong, Yuanjun and Yu, Stella X and Lin, Dahua},
	booktitle={IEEE Conference on Computer Vision and Pattern Recognition},
	pages={3733--3742},
	year={2018}
}

@article{wang2021exploring,
	title={Exploring cross-image pixel contrast for semantic segmentation},
	author={Wang, Wenguan and Zhou, Tianfei and Yu, Fisher and Dai, Jifeng and Konukoglu, Ender and Van Gool, Luc},
	journal={arXiv preprint arXiv:2101.11939},
	year={2021}
}

@article{zhao2020contrastive,
	title={Contrastive Learning for Label-Efficient Semantic Segmentation},
	author={Zhao, Xiangyun and Vemulapalli, Raviteja and Mansfield, Philip and Gong, Boqing and Green, Bradley and Shapira, Lior and Wu, Ying},
	journal={arXiv preprint arXiv:2012.06985},
	year={2020}
}

@inproceedings{NEURIPS2020_d89a66c7,
	author = {Khosla, Prannay and Teterwak, Piotr and Wang, Chen and Sarna, Aaron and Tian, Yonglong and Isola, Phillip and Maschinot, Aaron and Liu, Ce and Krishnan, Dilip},
	booktitle = {Advances in Neural Information Processing Systems},
	pages = {18661--18673},
	publisher = {Curran Associates, Inc.},
	title = {Supervised Contrastive Learning},
	volume = {33},
	year = {2020}
}

@article{tian2019contrastive,
	title={Contrastive multiview coding},
	author={Tian, Yonglong and Krishnan, Dilip and Isola, Phillip},
	journal={arXiv preprint arXiv:1906.05849},
	year={2019}
}

@article{zurn2020self,
	title={Self-supervised visual terrain classification from unsupervised acoustic feature learning},
	author={Z{\"u}rn, Jannik and Burgard, Wolfram and Valada, Abhinav},
	journal={Transactions on Robotics},
	year={2020},
	publisher={IEEE}
}

@INPROCEEDINGS{gaobiaoiv,
	author={Gao, Biao and Xu, Anran and Pan, Yancheng and Zhao, Xijun and Yao, Wen and Zhao, Huijing},
	booktitle={IEEE Intelligent Vehicles Symposium}, 
	title={Off-Road Drivable Area Extraction Using 3D LiDAR Data}, 
	year={2019},
	pages={1505-1511}
}

@inproceedings{tang2017one,
	title={From one to many: Unsupervised traversable area segmentation in off-road environment},
	author={Tang, Li and Ding, Xiaqing and Yin, Huan and Wang, Yue and Xiong, Rong},
	booktitle={2017 IEEE International Conference on Robotics and Biomimetics (ROBIO)},
	pages={787--792},
	year={2017},
	organization={IEEE}
}

@article{sharma2019semantic,
	title={Semantic segmentation with transfer learning for off-road autonomous driving},
	author={Sharma, Suvash and Ball, John E and Tang, Bo and Carruth, Daniel W and Doude, Matthew and Islam, Muhammad Aminul},
	journal={Sensors},
	volume={19},
	number={11},
	pages={2577},
	year={2019},
	publisher={Multidisciplinary Digital Publishing Institute}
}

@inproceedings{kim2018season,
	title={Season-invariant semantic segmentation with a deep multimodal network},
	author={Kim, Dong-Ki and Maturana, Daniel and Uenoyama, Masashi and Scherer, Sebastian},
	booktitle={Field and Service Robotics},
	pages={255--270},
	year={2018},
	organization={Springer}
}

@conference{Maturana17,
	author = {Daniel Maturana and Po-Wei Chou and Masashi Uenoyama and Sebastian Scherer},
	title = {Real-time Semantic Mapping for Autonomous Off-Road Navigation},
	booktitle = {International Conference on Field and Service Robotics},
	year = {2018},
	pages = {335--350}
}

@inproceedings{chiodini2020evaluation,
	title={Evaluation of 3D CNN Semantic Mapping for Rover Navigation},
	author={Chiodini, Sebastiano and Torresin, Luca and Pertile, Marco and Debei, Stefano},
	booktitle={IEEE International Workshop on Metrology for AeroSpace},
	pages={32--36},
	year={2020},
	organization={IEEE}
}

@inproceedings{viswanath2021offseg,
	title={Offseg: A semantic segmentation framework for off-road driving},
	author={Viswanath, Kasi and Singh, Kartikeya and Jiang, Peng and Sujit, PB and Saripalli, Srikanth},
	booktitle={IEEE International Conference on Automation Science and Engineering},
	pages={354--359},
	year={2021},
	organization={IEEE}
}

@article{guan2021ganav,
	title={Ganav: Group-wise attention network for classifying navigable regions in unstructured outdoor environments},
	author={Guan, Tianrui and Kothandaraman, Divya and Chandra, Rohan and Sathyamoorthy, Adarsh Jagan and Manocha, Dinesh},
	journal={arXiv preprint arXiv:2103.04233},
	year={2021}
}

@article{jin2021memory,
	title={Memory-based Semantic Segmentation for Off-road Unstructured Natural Environments},
	author={Jin, Youngsaeng and Han, David K and Ko, Hanseok},
	journal={arXiv preprint arXiv:2108.05635},
	year={2021}
}

@article{sgibnev2020deep,
	title={Deep semantic segmentation for the off-road autonomous driving},
	author={Sgibnev, I and Sorokin, A and Vishnyakov, B and Vizilter, Y},
	journal={The International Archives of Photogrammetry, Remote Sensing and Spatial Information Sciences},
	volume={43},
	pages={617--622},
	year={2020},
	publisher={Copernicus GmbH}
}

@incollection{rothrock2016spoc,
	title={Spoc: Deep learning-based terrain classification for mars rover missions},
	author={Rothrock, Brandon and Kennedy, Ryan and Cunningham, Chris and Papon, Jeremie and Heverly, Matthew and Ono, Masahiro},
	booktitle={AIAA SPACE},
	pages={5539},
	year={2016}
}

@inproceedings{zhou2010self,
	title={Self-supervised learning method for unstructured road detection using fuzzy support vector machines},
	author={Zhou, Shengyan and Iagnemma, Karl},
	booktitle={International Conference on Intelligent Robots and Systems},
	pages={1183--1189},
	year={2010},
	organization={IEEE}
}

@article{jeong2002vision,
	title={Vision-based adaptive and recursive tracking of unpaved roads},
	author={Jeong, Hong and Oh, Yuns and Park, Jeong-Ho and Koo, BS and Lee, Sang Wook},
	journal={Pattern Recognition Letters},
	volume={23},
	number={1-3},
	pages={73--82},
	year={2002},
	publisher={Elsevier}
}

@article{mei2017scene,
	title={Scene-adaptive off-road detection using a monocular camera},
	author={Mei, Jilin and Yu, Yufeng and Zhao, Huijing and Zha, Hongbin},
	journal={Transactions on Intelligent Transportation Systems},
	volume={19},
	number={1},
	pages={242--253},
	year={2017},
	publisher={IEEE}
}

@article{settles2009active,
	title={Active learning literature survey},
	author={Settles, Burr},
	year={2009},
	publisher={University of Wisconsin-Madison Department of Computer Sciences}
}

@article{oord2018CPC,
	title={Representation learning with contrastive predictive coding},
	author={Oord, Aaron van den and Li, Yazhe and Vinyals, Oriol},
	journal={arXiv preprint arXiv:1807.03748},
	year={2018}
}

@inproceedings{he2020momentum,
	title={Momentum contrast for unsupervised visual representation learning},
	author={He, Kaiming and Fan, Haoqi and Wu, Yuxin and Xie, Saining and Girshick, Ross},
	booktitle={Conference on Computer Vision and Pattern Recognition},
	pages={9729--9738},
	year={2020}
}

@inproceedings{holder2016road,
	title={From on-road to off: transfer learning within a deep convolutional neural network for segmentation and classification of off-road scenes},
	author={Holder, Christopher J and Breckon, Toby P and Wei, Xiong},
	booktitle={European Conference on Computer Vision},
	pages={149--162},
	year={2016},
	organization={Springer}
}

@inproceedings{wigness2019rugd,
	title={A rugd dataset for autonomous navigation and visual perception in unstructured outdoor environments},
	author={Wigness, Maggie and Eum, Sungmin and Rogers, John G and Han, David and Kwon, Heesung},
	booktitle={IEEE/RSJ International Conference on Intelligent Robots and Systems},
	pages={5000--5007},
	year={2019},
	organization={IEEE}
}

@inproceedings{behley2019semantickitti,
	title={{SemanticKITTI}: A dataset for semantic scene understanding of lidar sequences},
	author={Behley, Jens and Garbade, Martin and Milioto, Andres and Quenzel, Jan and Behnke, Sven and Stachniss, Cyrill and Gall, Jurgen},
	booktitle={IEEE International Conference on Computer Vision},
	pages={9297--9307},
	year={2019}
}

@inproceedings{cordts2016cityscapes,
	title={The cityscapes dataset for semantic urban scene understanding},
	author={Cordts, Marius and Omran, Mohamed and Ramos, Sebastian and Rehfeld, Timo and Enzweiler, Markus and Benenson, Rodrigo and Franke, Uwe and Roth, Stefan and Schiele, Bernt},
	booktitle={IEEE Conference on Computer Vision and Pattern Recognition},
	pages={3213--3223},
	year={2016}
}

@article{minaee2021image,
	title={Image segmentation using deep learning: A survey},
	author={Minaee, Shervin and Boykov, Yuri Y and Porikli, Fatih and Plaza, Antonio J and Kehtarnavaz, Nasser and Terzopoulos, Demetri},
	journal={IEEE Transactions on Pattern Analysis and Machine Intelligence},
	year={2021},
	publisher={IEEE}
}

@article{wellhausen2019should,
	title={Where should {I} walk? predicting terrain properties from images via self-supervised learning},
	author={Wellhausen, Lorenz and Dosovitskiy, Alexey and Ranftl, Ren{\'e} and Walas, Krzysztof and Cadena, Cesar and Hutter, Marco},
	journal={IEEE Robotics and Automation Letters},
	volume={4},
	number={2},
	pages={1509--1516},
	year={2019},
	publisher={IEEE}
}

@inproceedings{wang2009unstructured,
	title={Unstructured road detection using hybrid features},
	author={Wang, Jian and Ji, Zhong and Su, Yu-Ting},
	booktitle={International Conference on Machine Learning and Cybernetics},
	volume={1},
	pages={482--486},
	year={2009},
	organization={IEEE}
}

@inproceedings{alon2006off,
	title={Off-road path following using region classification and geometric projection constraints},
	author={Alon, Yaniv and Ferencz, Andras and Shashua, Amnon},
	booktitle={IEEE Conference on Computer Vision and Pattern Recognition},
	volume={1},
	pages={689--696},
	year={2006},
	organization={IEEE}
}

@article{shi2015fast,
	title={Fast and robust vanishing point detection for unstructured road following},
	author={Shi, Jinjin and Wang, Jinxiang and Fu, Fangfa},
	journal={IEEE Transactions on Intelligent Transportation Systems},
	volume={17},
	number={4},
	pages={970--979},
	year={2015},
	publisher={IEEE}
}

@inproceedings{kong2009vanishing,
	title={Vanishing point detection for road detection},
	author={Kong, Hui and Audibert, Jean-Yves and Ponce, Jean},
	booktitle={IEEE Conference on Computer Vision and Pattern Recognition},
	pages={96--103},
	year={2009},
	organization={IEEE}
}

@article{badue2021self,
	title={Self-driving cars: A survey},
	author={Badue, Claudine and Guidolini, R{\^a}nik and Carneiro, Raphael Vivacqua and Azevedo, Pedro and Cardoso, Vinicius B and Forechi, Avelino and Jesus, Luan and Berriel, Rodrigo and Paixao, Thiago M and Mutz, Filipe and others},
	journal={Expert Systems with Applications},
	volume={165},
	pages={113816},
	year={2021},
	publisher={Elsevier}
}

@inproceedings{shariati2019towards,
	title={Towards autonomous mining via intelligent excavators},
	author={Shariati, Hooman and Yeraliyev, Anuar and Terai, Burhan and Tafazoli, Shahram and Ramezani, Mahdi},
	booktitle={IEEE Conference on Computer Vision and Pattern Recognition Workshops},
	pages={26--32},
	year={2019}
}

@article{fang2006trajectory,
	title={Trajectory tracking control of farm vehicles in presence of sliding},
	author={Fang, Hao and Fan, Ruixia and Thuilot, Benoit and Martinet, Philippe},
	journal={Robotics and Autonomous Systems},
	volume={54},
	number={10},
	pages={828--839},
	year={2006},
	publisher={Elsevier}
}

@article{braid2006terramax,
	title={The TerraMax autonomous vehicle},
	author={Braid, Deborah and Broggi, Alberto and Schmiedel, Gary},
	journal={Journal of Field Robotics},
	volume={23},
	number={9},
	pages={693--708},
	year={2006},
	publisher={Wiley Online Library}
}

@article{badrinarayanan2017segnet,
	title={Segnet: A deep convolutional encoder-decoder architecture for image segmentation},
	author={Badrinarayanan, Vijay and Kendall, Alex and Cipolla, Roberto},
	journal={IEEE transactions on pattern analysis and machine intelligence},
	volume={39},
	number={12},
	pages={2481--2495},
	year={2017},
	publisher={IEEE}
}

@article{liu2015parsenet,
	title={Parsenet: Looking wider to see better},
	author={Liu, Wei and Rabinovich, Andrew and Berg, Alexander C},
	journal={arXiv preprint arXiv:1506.04579},
	year={2015}
}

@article{krahenbuhl2011efficient,
	title={Efficient inference in fully connected crfs with gaussian edge potentials},
	author={Kr{\"a}henb{\"u}hl, Philipp and Koltun, Vladlen},
	journal={Advances in neural information processing systems},
	volume={24},
	pages={109--117},
	year={2011}
}

@article{van2008visualizing,
	title={Visualizing data using t-SNE.},
	author={Van der Maaten, Laurens and Hinton, Geoffrey},
	journal={Journal of machine learning research},
	volume={9},
	number={11},
	year={2008}
}

@article{fraley1998algorithms,
	title={Algorithms for model-based Gaussian hierarchical clustering},
	author={Fraley, Chris},
	journal={SIAM Journal on Scientific Computing},
	volume={20},
	number={1},
	pages={270--281},
	year={1998},
	publisher={SIAM}
}

@InProceedings{Wu_2018_CVPR,
	author = {Wu, Zhirong and Xiong, Yuanjun and Yu, Stella X. and Lin, Dahua},
	title = {Unsupervised Feature Learning via Non-Parametric Instance Discrimination},
	booktitle = {Conference on Computer Vision and Pattern Recognition},
	month = {June},
	year = {2018}
}

@article{oord2018representation,
	title={Representation learning with contrastive predictive coding},
	author={Oord, Aaron van den and Li, Yazhe and Vinyals, Oriol},
	journal={arXiv preprint arXiv:1807.03748},
	year={2018}
}

@article{krizhevsky2012imagenet,
	title={{ImageNet} classification with deep convolutional neural networks},
	author={Krizhevsky, Alex and Sutskever, Ilya and Hinton, Geoffrey E},
	journal={Advances in Neural Information Processing Systems},
	volume={25},
	pages={1097--1105},
	year={2012}
}

@InProceedings{valada16iser,
	author = {Abhinav Valada and Gabriel Oliveira and Thomas Brox and Wolfram Burgard},
	title = {Deep Multispectral Semantic Scene Understanding of Forested Environments using Multimodal Fusion},
	booktitle = {International Symposium on Experimental Robotics (ISER)},
	year = {2016},
}

@inproceedings{long2015fully,
	title={Fully convolutional networks for semantic segmentation},
	author={Long, Jonathan and Shelhamer, Evan and Darrell, Trevor},
	booktitle={IEEE Conference on Computer Vision and Pattern Recognition},
	pages={3431--3440},
	year={2015}
}

@article{mclachlan2019finite,
	title={Finite mixture models},
	author={McLachlan, Geoffrey J and Lee, Sharon X and Rathnayake, Suren I},
	journal={Annual review of statistics and its application},
	volume={6},
	pages={355--378},
	year={2019},
	publisher={Annual Reviews}
}

@article{schwarz1978estimating,
	title={Estimating the dimension of a model},
	author={Schwarz, Gideon},
	journal={The annals of statistics},
	pages={461--464},
	year={1978},
	publisher={JSTOR}
}

\begin{IEEEbiography}
	[{\includegraphics[width=1.2in,height=1.2in,clip,keepaspectratio]{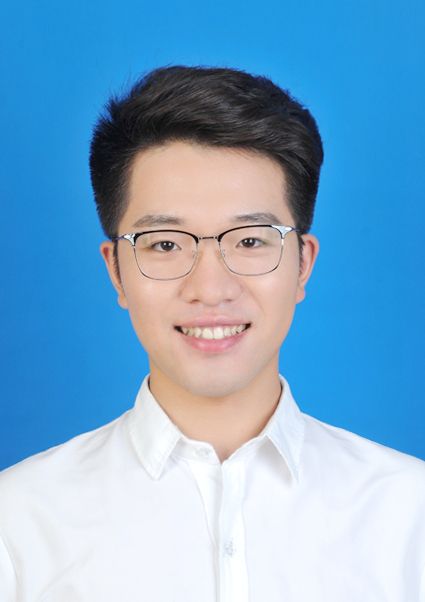}}]{Biao Gao}
	received B.S. degree in computer science (machine intelligence) from Peking University, Beijing, China, in 2017, where he is currently pursuing the Ph.D. degree with the Key Laboratory of Machine Perception (MOE), Peking University.
	His research interests include intelligent vehicles, 3D LiDAR perception, computer vision, and deep learning.
\end{IEEEbiography}
\begin{IEEEbiography}
[{\includegraphics[width=1.2in,height=1.2in,clip,keepaspectratio]{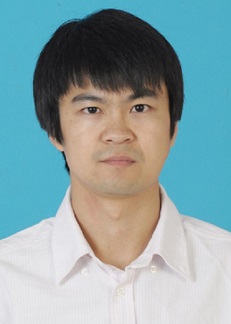}}]{Xijun Zhao}
 was born in Yanji City, Jilin, China, in 1984. He received the B.S degrees in vehicular engineering from Beijing Institute of Technology, Beijing, China, in 2007 and the Ph.D degree in mechanical engineering from Beijing Institute of Technology, Beijing, China, in 2011.
He is now working with China North Vehicle Research Institute. His research interests include perception, localization, motion planning and control of Unmanned Ground Vehicles.
\end{IEEEbiography}
\begin{IEEEbiography}
	[{\includegraphics[width=1.2in,height=1.2in,clip,keepaspectratio]{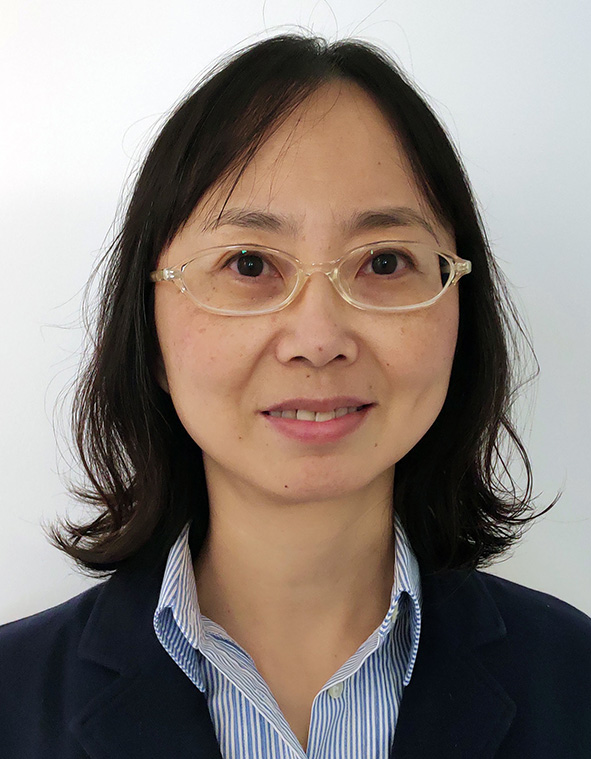}}]{Huijing Zhao}
	received B.S. degree in computer science from Peking University in 1991. She obtained M.E. degree in 1996 and Ph.D. degree in 1999 in civil engineering from the University of Tokyo, Japan. From 1999 to 2007, she was a postdoctoral researcher and visiting associate professor at the Center for Space Information Science, University of Tokyo. In 2007, she joined Peking University as a tenure-track professor at the School of Electronics Engineering and Computer Science. She became an associate professor with tenure on 2013 and was promoted to full professor on 2020. She has research interest in several areas in connection with intelligent vehicle and mobile robot, such as machine perception, behavior learning and motion planning, and she has special interests on the studies through real world data collection.
	
\end{IEEEbiography}

\end{document}